\documentclass{article}
\usepackage[preprint]{colm2026_conference}

\usepackage{microtype}
\usepackage{hyperref}
\usepackage{url}
\usepackage{booktabs}

\usepackage{lineno}

\usepackage[T1]{fontenc}
\usepackage[utf8]{inputenc}
\usepackage{inconsolata}

\usepackage{graphicx}

\usepackage{array}
\usepackage{tabularx}
\usepackage{amsmath}
\usepackage{multirow}
\usepackage{xcolor}
\usepackage{subcaption}
\usepackage{wrapfig}
\usepackage{listings}
\lstdefinestyle{promptstyle}{
  basicstyle=\scriptsize\ttfamily,
  breaklines=true,
  breakatwhitespace=true,
  columns=fullflexible,
  frame=single,
  framesep=4pt,
  xleftmargin=6pt,
  xrightmargin=6pt,
  aboveskip=6pt,
  belowskip=6pt,
  showstringspaces=false,
  keepspaces=true,
}
\lstnewenvironment{promptbox}{\lstset{style=promptstyle}}{}

\definecolor{darkblue}{rgb}{0, 0, 0.5}
\hypersetup{colorlinks=true, citecolor=darkblue, linkcolor=darkblue, urlcolor=darkblue}

\title{The Task Is Latent: Do Language Models Align with Users on Their Tasks?}
\title{Interactive Task Alignment as a POMDP}

\author{
    Andy Dai\textsuperscript{*,1} \quad
    Zexue He\textsuperscript{*,1} \quad
    Zhenyu Zhang\textsuperscript{1} \quad
    Alex Pentland\textsuperscript{1} \quad
    Jiaxin Pei\textsuperscript{1} \\
    $^{1}$Stanford University \\
    \texttt{aldai@cs.stanford.edu}
}

\begin{document}

\ifcolmsubmission
\linenumbers
\fi

\maketitle
\begingroup
\renewcommand{\thefootnote}{*}
\footnotetext{Equal contribution.}
\endgroup

\begin{abstract}

Current benchmarks for language models primarily evaluate execution on fully specified tasks. However, real user tasks are often ambiguous. Users arrive with incomplete, exploratory, or even inconsistent goals, requiring the assistant to first determine the intended task before carrying it out. We study this problem as \textit{task alignment}: the ability to align with a user on their intended task. We introduce a general framework for converting \textbf{specified tasks} into \textbf{underspecified interactions}, formalized as a POMDP in which the model must infer a latent task from partial and evolving user intent. We validate our user simulator post hoc with a human user study. Across shopping, coding, and professional work settings, we find that while models often perform well once the task is specified, models still struggle with task alignment: current models act prematurely, interact ineffectively, and fail to resolve ambiguous requests. Models on average recover the user's intended task only 22 -- 32\% of the time under ambiguity. In a human study in the same setting, humans reach 48\%, outperforming all evaluated models. We show that post-training with supervised fine-tuning and reinforcement learning improves task alignment, but models still lag behind humans in resolving uncertainty through interaction. Together, our results suggest that current models still lack key interaction abilities required for reliable agency. \footnote{See \href{https://huggingface.co/datasets/daiandy/task-alignment-dataset}{data} and \href{https://github.com/dai-andy/task-alignment}{code}.}

\end{abstract}

\section{Introduction}
\label{sec:introduction}
Large language models (LLMs) are primarily evaluated on fully specified tasks. The model is given a goal, constraints, and success criteria, and is evaluated on whether it can execute the task \citep{yao2023webshopscalablerealworldweb}, \citep{wei2025browsecompsimplechallengingbenchmark}, \citep{jimenez2024swebenchlanguagemodelsresolve}. This fully specified setting is useful for isolating \emph{task capability}. For example, to measure a model's ability to write accurate code, one must fully specify the coding issue and its tests, so that performance reflects the model's actual code-writing capability, rather than its ability to interpret requests \citep{jimenez2024swebenchlanguagemodelsresolve}.  

In real world deployments, however, users almost never begin with complete specifications. Humans typically express goals that are incomplete, exploratory, ambiguous, or even inconsistent over time. As a result, the bottleneck shifts from \emph{executing a known task} to \emph{determining which task the user actually intends}.

\begin{wrapfigure}{r}{0.35\linewidth}
  \vspace{-22pt}
  \centering
    \includegraphics[width=\linewidth]{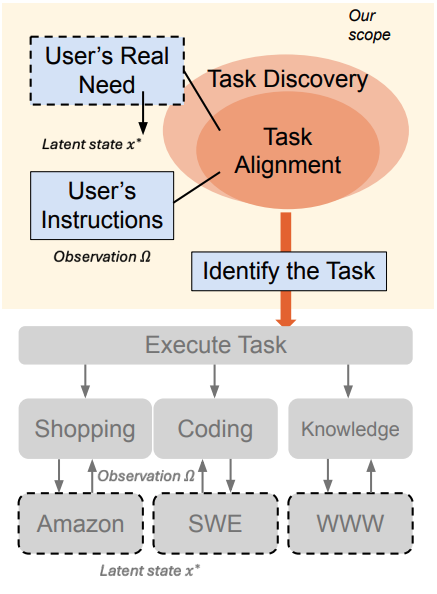}
    \caption{Task alignment under uncertainty. Users provide observable instructions, while their real needs remain latent. We study the task alignment stage: identifying the intended task through interaction before executing the task. Although task execution may involve additional latent environment states, these downstream execution dynamics are out of scope.}
  \vspace{-22pt}
  \label{fig:concept}
\end{wrapfigure}

Recent work has begun to recognize this discrepancy between fully specified evaluation and real-world use. Examples include 
Ambig-SWE \citep{vijayvargiya2026ambigsweinteractiveagentsovercome} on software engineering tasks and GDPval \citep{patwardhan2025gdpvalevaluatingaimodel} on economically valuable, real-world tasks. Such benchmarks are important steps toward realistic evaluation under task ambiguity, but they typically measure end-to-end task completion with underspecification. This makes it difficult to separate failures of task execution from failures of task interpretation. In this paper, we decouple the two and study the model's ability to achieve \textbf{task alignment} \footnote{Task alignment: the process by which agents ground and maintain a shared task understanding through mutual modeling \citet{clark1991grounding}} with users (Figure~\ref{fig:concept}). We argue that task alignment is a prerequisite for downstream task completion: even highly capable models can fail when they fail to detect missing information, assume the user's intent, and act prematurely.

To this end, we propose a domain-general framework for turning fully specified task evaluations into task alignment evaluations with underspecified user interactions. Given a domain with a space of tasks and specifications, we construct an interactive setting in which the ground-truth task specification is latent, partially revealed through a user simulator, and progressively evolved from abstract to fully specified. We formalize this setting as a \emph{Partially Observable Markov Decision Process} (POMDP), where the assistant observes only the dialogue history and must interact with the user, update its belief over possible tasks, and decide when to commit to task execution. This formulation separates task discovery from task execution and makes the framework applicable beyond any single domain.

We apply \emph{behavioral} and \emph{information-theoretic} metrics of the interaction process to evaluate task alignment under ambiguity, along the following two complementary dimensions:  (1) \emph{task recovery} measures whether the model identifies the intended task; (2) \textit{uncertainty resolution} measures whether the model appropriately detects ambiguity, seeks informative clarifications, and reduces uncertainty through interaction.

We conduct a detailed evaluation of state-of-the-art LLMs across real-world shopping, coding, and professional work tasks. Our analysis shows that current models struggle at aligning with users on their tasks, recovering on average the correct task between $22-32\%$ of the time for underspecified users. Models that perform well under full specification often act prematurely, committing to task execution in 1-3 conversation turns; interact ineffectively, failing to clarify the task; or remain uncertain about the user's intended task.
 Our results suggest that current LLMs still lack key communicative behaviors required for reliable agency. 
 
 Key contributions from this paper include: (1) we propose a framework formalized as a POMDP for measuring how well models align with users on any specifiable task, and we validate the framework with a human study, showing simulated users elicit human-comparable specification behavior; (2) we show that current models struggle at determining users' tasks even when they can execute them, and that task alignment is a separable, domain-general axis of model behavior; (3) we demonstrate that humans exceed all evaluated models at identifying underspecified users' tasks; (4) we show that post-training Qwen3.5 9B with SFT and RL improves task alignment, but models still lag behind human performance.

\section{Problem Formulation}

\subsection{A POMDP Framework}
\label{subsec:framework}

\begin{figure}
  \centering
\includegraphics[width=\linewidth]{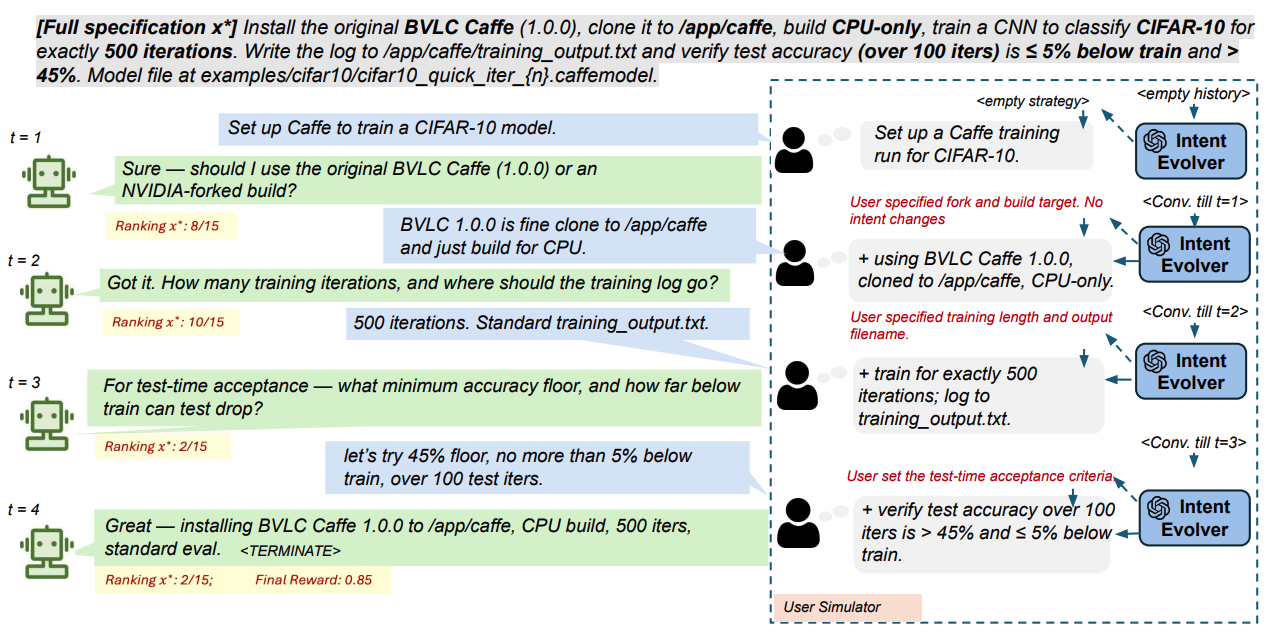}
  \caption{Workflow of our POMDP-based task-alignment framework, illustrated with an example from Terminal-Bench. The assistant interacts with a user simulator whose responses are guided by an evolving intent state conditioned on the full specification and conversation history. At each turn, the assistant ranks candidate specifications (as an additional output just for evaluation), allowing us to track how interaction affects belief over the latent task.}
  \label{fig:overview}
\end{figure}

Given a task setting $T$, let $\mathcal{X}_T$ be the set of all possible task specifications induced by $T$. For example, in shopping setting, a task specification could be buying a particular Amazon product (``\textit{purchase a Bose QuietComfort Bluetooth Headphones}''): in coding, it could be a problem statement (``\textit{Build POV-Ray 2.2. Find and download the source...}'') \citep{merrill2026terminalbenchbenchmarkingagentshard}.

We formulate \textbf{task alignment} as a Partially Observable Markov Decision Process (POMDP) $\mathcal{M}_T = \langle\, \mathcal{S},\, \mathcal{A},\, \mathcal{T}, \, R,\, \Omega,\, \mathcal{O} \,\rangle$ from the assistant's perspective as follows.

\paragraph{State $\mathcal{S}$ and observation $\Omega$.}
At turn $t$, the environment state is $s_t = (x^*, i_t, h_t) \in \mathcal{S}$, where $x^* \in \mathcal{X}_T$ is the fixed ground-truth specification, $i_t \in \mathcal{I}$ is the user's current intent state, and $h_t=(m_1^u,m_1^a,\ldots,m_{t-1}^u,m_{t-1}^a)$ is the conversation history before the assistant commits to execution and terminates the conversation. The intent state captures which aspects of the user's goal have been revealed, refined, or left implicit. The assistant observes only the conversation history, so $\Omega$ is the space of possible conversations and the observation function is deterministic ($o_t = h_t$). Both $x^*$ and $i_t$ are hidden from the assistant, and must therefore be inferred from the interaction.

\paragraph{Action $\mathcal{A}$.}
At each turn, the assistant takes an action $a_t = (m_t^a, e_t, \rho_t) \in \mathcal{A}$,
where $m_t^a$ is the natural-language response (e.g., a clarifying question)
appended to $h_t$, and $e_t \in \{0,1\}$ is the stop signal parsed from $m_t^a$
indicating commitment to task execution, which terminates the conversation.
The third component $\rho_t$ is a ranking over a finite candidate set
$\mathcal{C} \subseteq \mathcal{X}_T$ (with $x^* \in \mathcal{C}$) that we
elicit in a \textit{separate conversation thread} after each turn, conditioned on the conversation
so far. $\rho_t$ is strictly a measurement instrument: it exposes the
assistant's current hypothesis about the intended task without leaking the
candidates into the conversation.

\paragraph{Transition $\mathcal{T}$.}
The ground-truth specification $x^\star$ is fixed throughout the episode;
only $h_t$ and $i_t$ evolve. If $e_t = 1$, the episode terminates. Otherwise
the user simulator generates a response $m_t^u \sim \pi_u(\cdot \mid i_t, h_t, m_t^a)$
and evolves the intent state $i_{t+1} \sim E(\cdot \mid x^\star, i_t, h_t, m_t^a, m_t^u)$,
where both the user policy $\pi_u$ and the intent evolver $E$ are implemented
with LLMs. The user simulator thus defines $\mathcal{T}$ during the rollout: it has privileged access
to $x^\star$ and reveals information about it only through responses to the
assistant's actions.

Task alignment evaluates whether an assistant policy $\pi_a(a_t \mid h_t)$, which maps observed conversation histories to distributions over actions, can use interaction to concentrate its hypothesis on $x^*$ while avoiding premature commitment under uncertainty.

\subsection{Evaluation Metrics}
\label{subsec:metrics}

We design metrics to (1) evaluate if the LLM assistant can recover latent task (\textit{task recovery}), and (2) to measure how it uses interaction to reduce uncertainty before committing (\textit{uncertainty resolution}). After each turn, the assistant provides a ranking $\rho_t$ over the candidate set $\mathcal{C}$, which allows us to track how the ground-truth specification $x^\star$ moves through the assistant's hypotheses over time. We aim to analyze the compounding effect of accuracy with uncertainty (Figure~\ref{fig:entropy-rank-scatter}).

\paragraph{Task recovery behavior.}
The first question is whether the assistant has recovered the intended task by the time it commits to execution. Let $T^*$ denote the termination turn. We report
\textbf{CommitAcc@1} $= \mathbb{1}\!\left[\mathrm{rank}_{\rho_{T^*}}(x^*) = 1\right]$;
the normalized rank $\mathbf{CommitRank} = \hat r_t = 1 - \frac{\mathrm{rank}_{\rho_t}(x^*) - 1}{|\mathcal{C}| - 1} \in [0,1]$,
focusing on $\hat r_1$ and $\hat r_{T^*}$; and
\textbf{CommitReward} $= \alpha \hat r_{T^*} - \lambda T^*$ with $\alpha = 1.5$
and $\lambda = 0.02$, which we use as the training signal in
Section~\ref{section:model-training} under different hyperparameters. Together with
$\mathbb{E}[T^*]$, these capture the quality-cost tradeoff at commitment: a well-aligned assistant should identify the correct specification while avoiding unnecessary interaction. 

\paragraph{Uncertainty resolution.}
Ranking behavior alone does not reveal whether the assistant used the conversation effectively. When the assistant emits confidence scores over $\mathcal{C}$ alongside $\rho_t$, we treat the normalized score vector as an empirical belief over candidates. Details are in Appendix~\ref{section:entropy-calculation}; uncertainty metrics are reported on the model subset that emits confidence scores. Let $\tilde{H}_t(\pi_a) \in [0,1]$ denote its normalized entropy at turn $t$. We summarize how much uncertainty the assistant resolves over the episode by the cumulative information gain:
\[
\boldsymbol{\mathrm{TotalIG}}(\pi_a)=\tilde{H}_0(\pi_a)-\tilde{H}_{T^*}(\pi_a).
\]
These quantities distinguish assistants that merely guess correctly from assistants that systematically narrow the space of plausible tasks through useful interaction. Crossing correctness with certainty at commitment yields four outcomes (Table~\ref{tab:taxonomy}), which we use throughout to separate assistants that are \textit{honestly uncertain} from those that are \textit{confidently wrong}.

\begin{table}[h]
\centering
\small
\begin{tabular}{@{}lcc@{}}
\toprule
 & \textbf{Low $\tilde{H}_{T^*}$ (confident)} & \textbf{High $\tilde{H}_{T^*}$ (uncertain)} \\ \midrule
$\mathrm{rank}_{\rho_{T^*}}(x^*) = 1$ (correct) & Confidently correct & Lucky correct \\
$\mathrm{rank}_{\rho_{T^*}}(x^*) > 1$ (incorrect) & Confidently wrong & Honestly uncertain \\ \bottomrule
\end{tabular}
\caption{Four-quadrant taxonomy of per-episode alignment outcomes at the assistant's commit turn $T^*$. Low and High $\tilde{H}_{T^*}$ is split by \textit{median} across all rollouts.}
\label{tab:taxonomy}
\end{table}

\subsection{Task Settings}
\label{subsec:task-settings}
Our framework includes three tasks: 
\textbf{Shopping-MMLU} \citep{jin2024shoppingmmlumassivemultitask}: purchase the specified product within a page of products;
\textbf{GDPVal }\citep{patwardhan2025gdpvalevaluatingaimodel}: produce the real-world deliverable that satisfies the rubric;
\textbf{Terminal-Bench }\citep{merrill2026terminalbenchbenchmarkingagentshard}: write code, modify files, or execute commands to pass the verifier. Converting a domain into a task alignment evaluation requires only its ground-truth task specifications $x^*$: from each $x^*$ we derive an initial user intent $i_0$, an intent evolver $E: (h_t, i_t, x^*) \mapsto i_{t+1}$ with privileged access to $x^*$, and a candidate set $\mathcal{C}$ with $|\mathcal{C}| = 15$ and $x^* \in \mathcal{C}$. The same recipe applies to any benchmark with specifiable tasks.
 
\paragraph{Initializing the intent state.}
We introduce underspecification by initializing $i_0$ at varying levels of abstraction. For each task, an LM rewrites $x^*$ as three intent tiers spoken from the user's perspective, ranging from \textit{abstract} (``I need something to keep him busy'') through \textit{moderate} (``I want a fun toy that'll hold his attention'') to \textit{concrete} (``I'm looking for something exciting for a little kid who loves big construction vehicles and gets bored easily with quiet toys'').\footnote{The ground-truth specification is ``purchase a Construction Excavator Toy Truck with Electric Universal Wheel for 3+ Years Old''.} The generated $i_0$ seeds the user simulator, and $E$ defines the sequence $\{i_t\}$ evolving from $i_0$ toward $x^*$.

\paragraph{Constructing the candidate set $\mathcal{C}$.}
All three tasks follow the same pipeline: we generate $N = 14$ perturbations of the ground-truth specification $x^*$, and insert $x^*$ at a random index to yield $|\mathcal{C}| = 15$. For Shopping, perturbations exist naturally as products similar to $x^*$: we extract the search term behind $x^*$ with an LM and take the first 14 Amazon.com search results. For GDPVal and Terminal-Bench, an LM generates perturbations along key dimensions (workflow, data inputs, deliverable, or methodology) and applies them to $x^*$. Full details and per-benchmark examples are in Appendix~\ref{section:task-setting-creation}.

\section{Experiments}
\subsection{Setups} 

\label{subsec:setups}

\begin{wrapfigure}[33]{r}{0.6\linewidth}
  \vspace{-22pt}
  \centering
    \includegraphics[width=\linewidth]{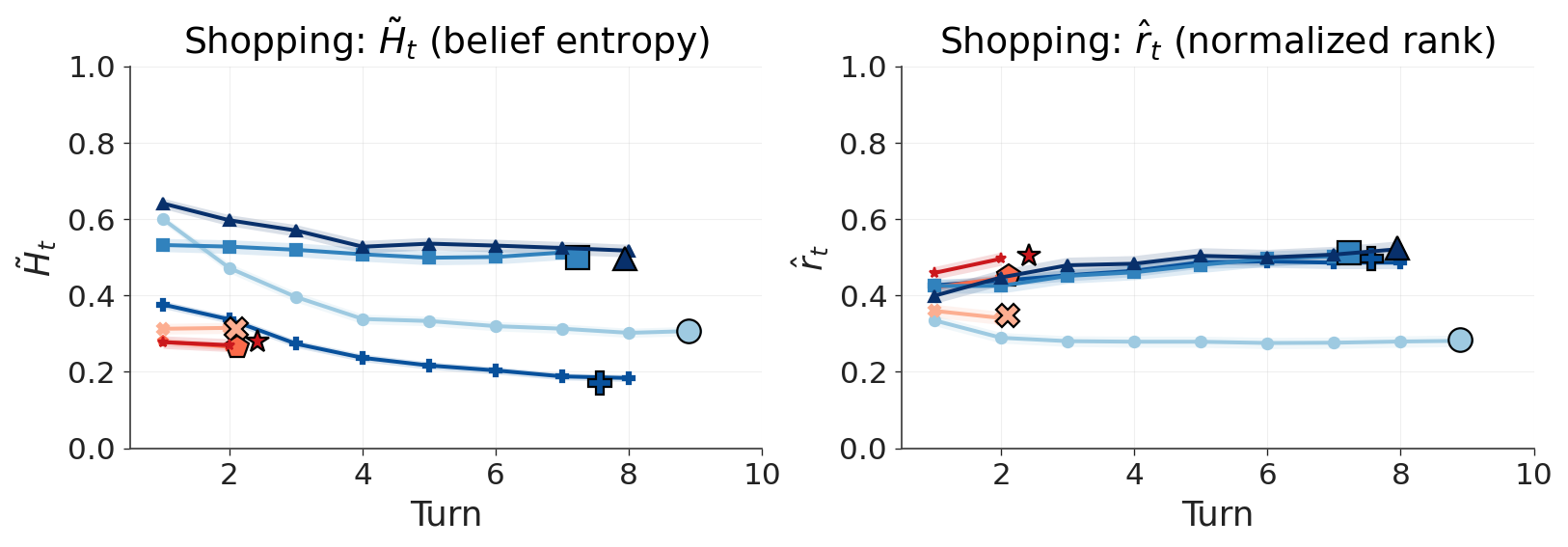}
    \includegraphics[width=\linewidth]{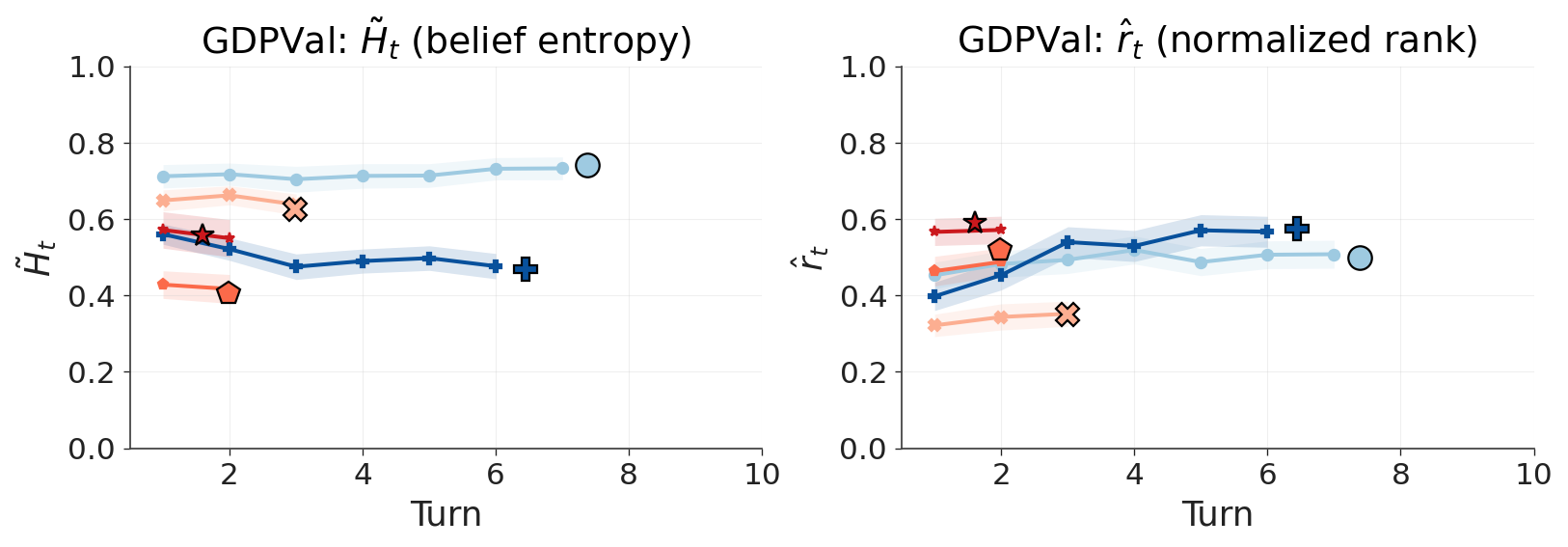}
    \includegraphics[width=\linewidth]{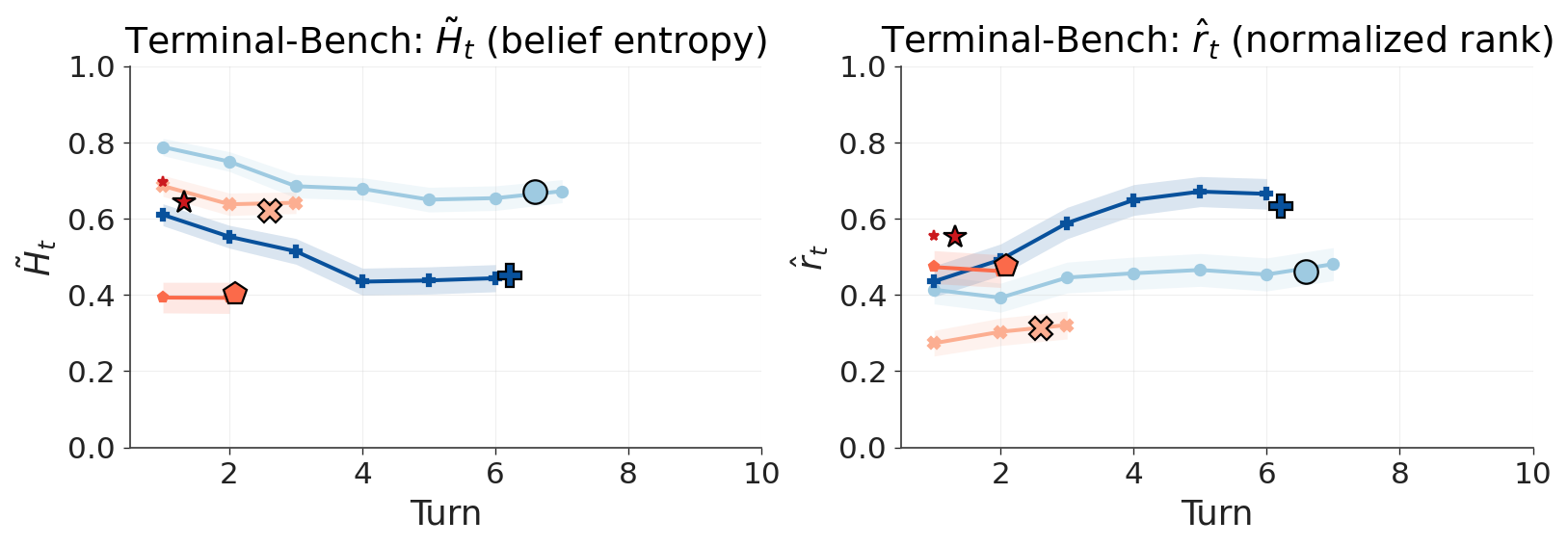}
    \includegraphics[width=0.85\linewidth]{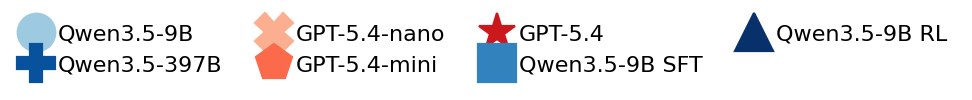}
  \caption{Per-turn trajectories of belief entropy $\tilde H_t$ (left) and normalized rank $\hat r_t$ (right) at the abstract tier, one row per task. Score-vector subset (GPT-5.4 family, Qwen3.5-9B, Qwen3.5-397B); trained Qwen3.5-9B SFT / RL added on Shopping. Trajectory lines terminate at mean $\lceil T^*\rceil$, and already-terminated rollouts contribute their terminal value thereafter for icon placement. GPT-5.4 family trajectories end early across all tasks due to \textit{early commitment} to task execution ($T^* \in [1, 2]$).
  }
  \vspace{-22pt}
  \label{fig:behavioral-decomp}
\end{wrapfigure}

We test current state-of-the-art closed-source LLMs with parameter sizes such as OpenAI GPT models including \texttt{GPT-5.4-\{nano, mini\}, GPT-5.4}, Claude \texttt{Sonnet-4.6}, and Google \texttt{Gemini-3-Flash}, \texttt{Gemini-3.1-Pro}. We also include open-source LLMs including Qwen3.5 series (\texttt{Qwen3.5-\{9B, 27B, 35B, 122B, 397B\}}), \texttt{Kimi-K2.5}, \texttt{Deepseek V3.2}, and \texttt{GLM-5.1}. All models are queried by official API call, and reasoning setting is set to \texttt{None} or lowest possible as applicable.

\paragraph{Task capability baseline.} We first establish each model's underlying task capability in the original task setting. We re-run our own evaluation on Shopping-MMLU and reuse normalized scores from the Artificial Analysis leaderboard for GDPVal and Terminal-Bench (per-model scores in Appendix~\ref{section:full-metrics}, Figure~\ref{fig:zero-shot}). Capability is saturated and highly clustered on Shopping-MMLU (scores in [81.1\%, 88.4\%] across all 14 models), but spans a wide range on GDPVal (52.8\%–100\%) and Terminal-Bench (19.7\%–100\%). We exploit this contrast in Section~\ref{subsec:cap-vs-align}: Shopping holds capability fixed as a control, while the two unsaturated settings let us study how capability and alignment relate.

\subsection{Task recovery behavior}
\label{subsec:task-recovery}

We first evaluate whether models can determine the correct task under underspecification. Table~\ref{tab:multitask-abstract} reports CommitReward and CommitAcc@1 on the abstract tier of all three tasks, while full metrics Tables~\ref{tab:drift-shopping} through \ref{tab:drift-terminalbench} are in the Appendix~\ref{section:full-metrics}.

\begin{table}[!htbp]
  \centering
  \tiny
  \caption{Task-alignment performance under abstract intent (14 models, all three tasks). CR = CommitReward; @1 = CommitAcc@1 (\%). Bold = per-column best within each task. AVERAGE = unweighted mean across the 14 models in this table.}
  \label{tab:multitask-abstract}
  \setlength{\tabcolsep}{2.5pt}
  \resizebox{\linewidth}{!}{
  \begin{tabular}{l rr rr rr}
    \toprule
    & \multicolumn{2}{c}{Shopping} & \multicolumn{2}{c}{GDPVal} & \multicolumn{2}{c}{Terminal-Bench} \\
    \cmidrule(lr){2-3}\cmidrule(lr){4-5}\cmidrule(lr){6-7}
    Model & CR & @1 (\%) & CR & @1 (\%) & CR & @1 (\%) \\
    \midrule
    qwen3.5-9b & $0.25 \pm 0.02$ & $15.80 \pm 1.31$ & $0.60 \pm 0.06$ & $25.26 \pm 4.48$ & $0.56 \pm 0.06$ & $25.00 \pm 4.75$ \\
    qwen3.5-27b & $0.68 \pm 0.02$ & $32.26 \pm 1.68$ & $0.67 \pm 0.06$ & $33.68 \pm 4.87$ & $0.73 \pm 0.06$ & $32.53 \pm 5.17$ \\
    qwen3.5-35b-a3b & $\mathbf{0.81 \pm 0.02}$ & $\mathbf{53.86 \pm 1.79}$ & $0.63 \pm 0.05$ & $27.37 \pm 4.60$ & $0.64 \pm 0.06$ & $29.27 \pm 5.06$ \\
    qwen3.5-122b-a10b & $0.67 \pm 0.02$ & $32.13 \pm 1.68$ & $0.62 \pm 0.06$ & $25.26 \pm 4.48$ & $0.64 \pm 0.07$ & $29.41 \pm 4.97$ \\
    qwen3.5-397b-a17b & $0.59 \pm 0.02$ & $23.87 \pm 1.53$ & $0.74 \pm 0.06$ & $38.95 \pm 5.03$ & $0.83 \pm 0.06$ & $39.29 \pm 5.36$ \\
    \cmidrule(lr){1-7}
    gpt-5.4-nano & $0.48 \pm 0.03$ & $7.97 \pm 1.37$ & $0.47 \pm 0.05$ & $9.47 \pm 3.02$ & $0.42 \pm 0.06$ & $7.06 \pm 2.79$ \\
    gpt-5.4-mini & $0.63 \pm 0.02$ & $11.44 \pm 1.14$ & $0.74 \pm 0.06$ & $17.89 \pm 3.95$ & $0.67 \pm 0.06$ & $17.65 \pm 4.16$ \\
    gpt-5.4 & $0.71 \pm 0.03$ & $12.89 \pm 1.70$ & $0.85 \pm 0.06$ & $22.11 \pm 4.28$ & $0.80 \pm 0.06$ & $24.71 \pm 4.71$ \\
    \cmidrule(lr){1-7}
    gemini-3-flash-preview & $0.71 \pm 0.02$ & $18.51 \pm 1.39$ & $0.87 \pm 0.06$ & $33.68 \pm 4.87$ & $0.81 \pm 0.06$ & $34.12 \pm 5.17$ \\
    gemini-3.1-pro-preview & $0.64 \pm 0.03$ & $13.88 \pm 1.76$ & $\mathbf{0.92 \pm 0.05}$ & $\mathbf{40.00 \pm 5.05}$ & $\mathbf{1.12 \pm 0.05}$ & $\mathbf{61.18 \pm 5.32}$ \\
    \cmidrule(lr){1-7}
    claude-sonnet-4-6 & $0.62 \pm 0.03$ & $12.60 \pm 1.68$ & $0.90 \pm 0.05$ & $31.58 \pm 4.79$ & $0.87 \pm 0.06$ & $34.12 \pm 5.17$ \\
    \cmidrule(lr){1-7}
    glm-5.1 & $0.77 \pm 0.02$ & $23.91 \pm 1.53$ & $0.77 \pm 0.05$ & $21.05 \pm 4.20$ & $0.98 \pm 0.06$ & $51.76 \pm 5.45$ \\
    deepseek-v3.2 & $0.78 \pm 0.02$ & $32.52 \pm 1.68$ & $0.75 \pm 0.06$ & $31.58 \pm 4.79$ & $0.80 \pm 0.06$ & $34.94 \pm 5.27$ \\
    kimi-k2.5 & $0.67 \pm 0.02$ & $16.45 \pm 1.33$ & $0.91 \pm 0.05$ & $28.42 \pm 4.65$ & $0.80 \pm 0.06$ & $29.41 \pm 4.97$ \\
    \midrule
    AVERAGE & $0.64 \pm 0.01$ & $22.01 \pm 0.42$ & $0.75 \pm 0.01$ & $27.59 \pm 1.21$ & $0.76 \pm 0.02$ & $32.17 \pm 1.32$ \\
    \bottomrule
  \end{tabular}
  }
\end{table}

\paragraph{Models struggle at aligning on real user tasks.}  We observe a large alignment gap between the tasks that models execute and the tasks that users intend. In the abstract tier, CommitAcc@1 collapses to averages of $22.01 \pm 0.42\%$ on Shopping, $27.59 \pm 1.21\%$ on GDPVal, and $32.17 \pm 1.32\%$ on Terminal-Bench. Table~\ref{tab:multitask-abstract} reports CommitReward and CommitAcc@1 with top performers in bold: Qwen3.5-35B-A3B leads Shopping on both metrics, while Gemini-3.1 Pro tops GDPVal and Terminal-Bench on both metrics. No model is able to consistently align with underspecified users across task settings, even after 15 turns of interaction.

\paragraph{Commit behavior is a stable model signature.} 
The timing for when each model commits to task execution is highly stable across all three domains, essentially independent of domain (Figure~\ref{fig:behavioral-decomp}; Figure~\ref{fig:tstar-full} in Appendix~\ref{section:full-metrics}; domain pairwise Spearman $\rho \geq 0.92$). GPT-5.4 commits within two turns everywhere ($T^\star \in [1.2, 2.4]$); Qwen3.5-9B takes many turns ($T^\star \in [6.4, 8.5]$); and Gemini and Claude families occupy the middle ($T^\star \in [3.7, 5.9]$).

\paragraph{Per-model performance is explained along three behavioral axes.}
We describe performance along: (1) the model's \textit{initial guess} $\hat r_1$ on the intended task, (2) how effectively the model leverages \textit{interaction} to determine the task $\Delta\hat r = \hat r_{T^\star} - \hat r_1$, and (3) \textit{when the model commits} $T^*$ to task execution. Figure~\ref{fig:behavioral-decomp} showcases a subset models' behavior along these axes; Figure~\ref{fig:behavioral-decomp-static} in Appendix~\ref{section:full-metrics} has full values. This decomposition reveals that (1) under abstract user intent, models' initial guesses for which task to execute are largely wrong; and (2) that interaction is necessary but insufficient for task alignment.

\subsection{Uncertainty resolution}

The previous focus was in describing whether the assistant could commit to the right task. Here, we analyze \textit{how} the model got there: how much of assistant belief entropy over $\mathcal{C}$ reduces over conversation. We report uncertainty metrics (Section~\ref{subsec:metrics}) for GPT-5.4 family and Qwen3.5-\{9B, 397B\}.

\begin{figure}[!htbp]
  \centering
  \includegraphics[width=0.32\linewidth]{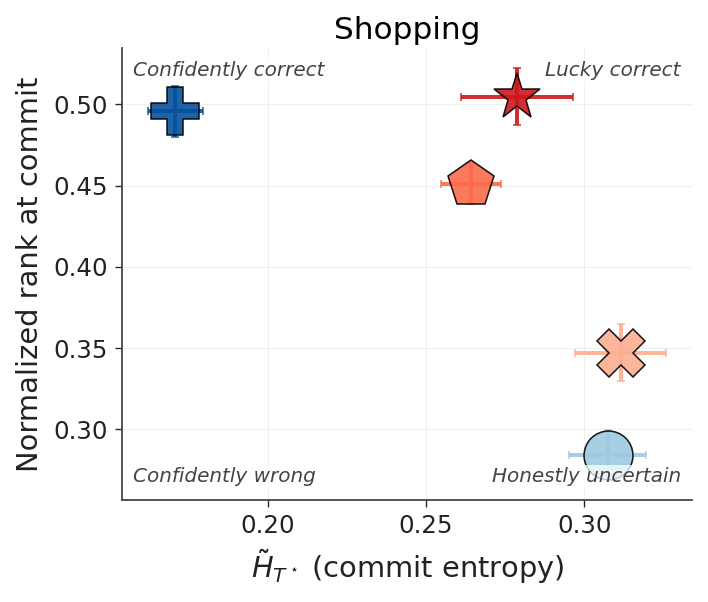}\hfill
  \includegraphics[width=0.32\linewidth]{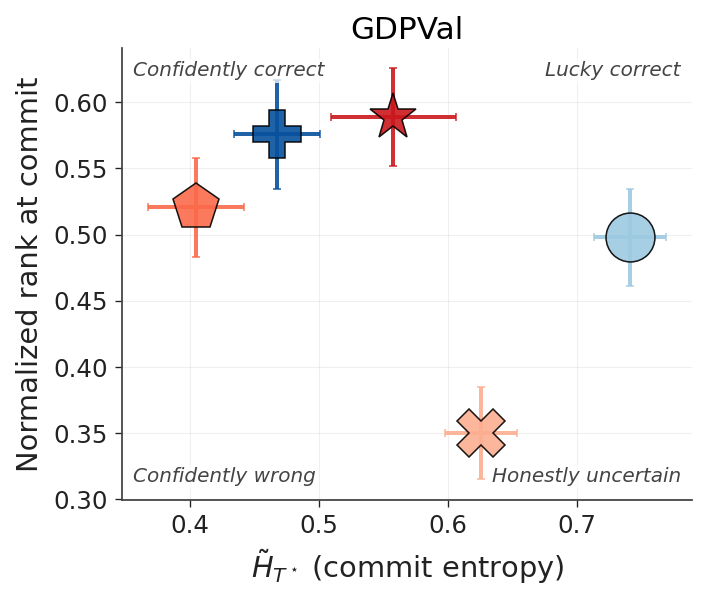}\hfill
  \includegraphics[width=0.32\linewidth]{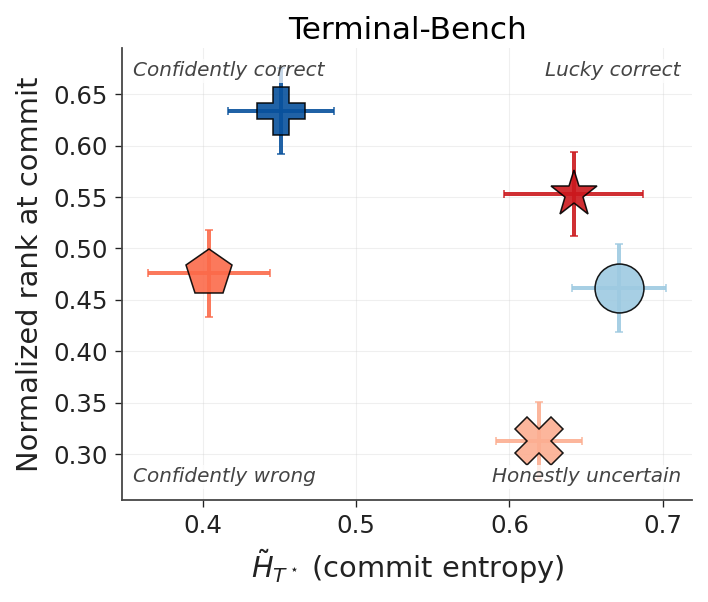}\\[6pt]
  \includegraphics[width=0.75\linewidth]{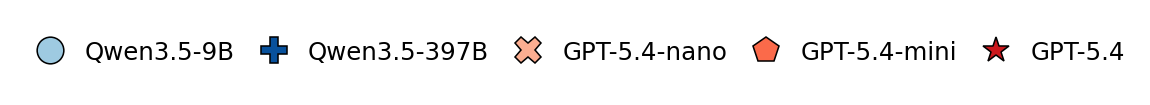}
  \caption{Normalized entropy $\tilde H$ at commit vs.\ normalized rank at commit, per model (mean $\pm$ SEM), across all three tasks.}
  \label{fig:entropy-rank-scatter}
\end{figure}

\paragraph{Qwen3.5 extracts more information from conversation than GPT-5.4}

Qwen3.5 achieves substantial entropy reductions on every task ($\mathrm{TotalIG}$ up to $0.29$ on Shopping and $0.21$ on Terminal-Bench), driven by more interaction turns ($T^* \in [6,8]$), while the GPT-5.4 family hovers around $\mathrm{TotalIG} = 0$ on Shopping, $0.03$ on GDPVal, and $0.02$ on Terminal-Bench due to low interaction ($T^* \in [1,3]$). Full per-(model, task) TotalIG is in Figure~\ref{fig:total-ig-barplot} in Appendix~\ref{section:full-metrics}.

\paragraph{Scaling model size improves rank, but confidence and calibration follow no consistent trend.} Within each family, CommitRank scales with model size on every task (Figure~\ref{fig:entropy-rank-scatter}), but confidence and calibration do not follow. Qwen3.5-397B carries the highest \emph{confidently correct} share ($21.1\%$ Shopping, $24.2\%$ GDPVal, $27.4\%$ Terminal-Bench), surpassing Qwen3.5 9B with lower entropy and higher rank: however, \emph{confidently wrong} also increases ($4.8 \to 17.9\%$ on Terminal-Bench, $7.4 \to 16.8\%$ on GDPVal). GPT-5.4 is likewise more accurate but not necessarily more certain than its smaller variants, as entropy traces a V-shape with GPT-5.4-mini at minimum entropy across all tasks (full values in Table~\ref{tab:quadrants}, Appendix~\ref{section:full-metrics}).

\subsection{Relationship between task capability and alignment}
\label{subsec:cap-vs-align}

\begin{wrapfigure}[17]{r}{0.46\linewidth}
  \vspace{-20pt}
  \centering
    \centering
    \includegraphics[width=\linewidth]{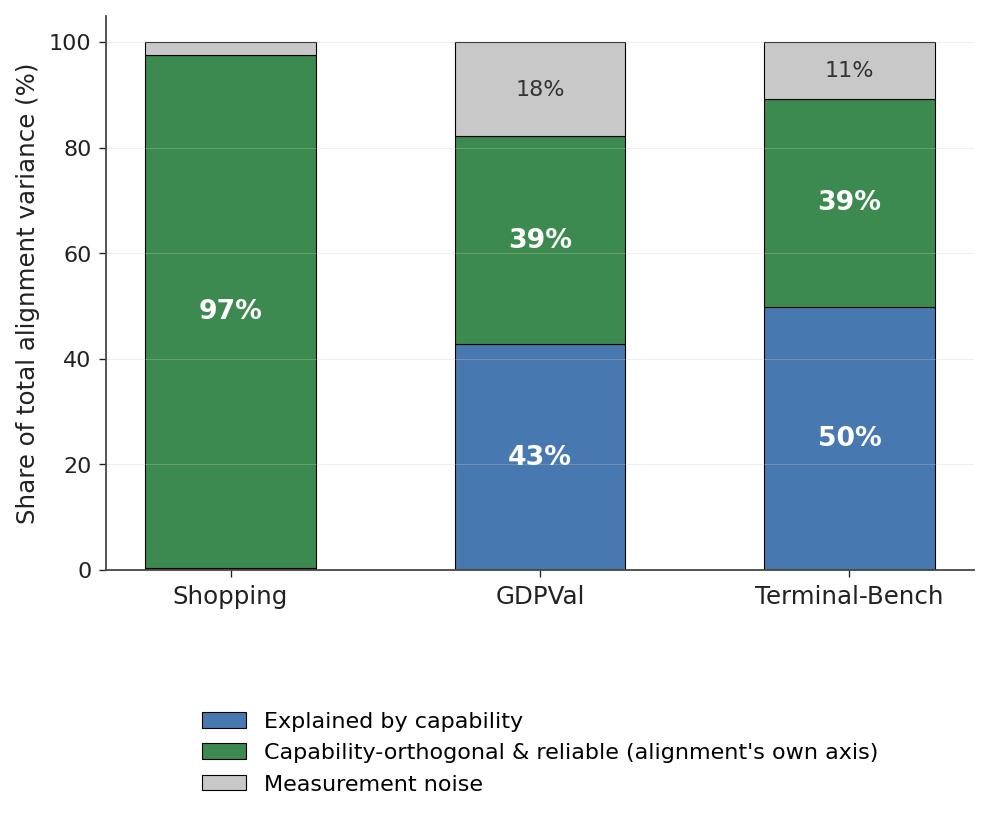}
    
  \caption{Variance decomposition of across-model alignment (CommitRank) per task.}
  \label{fig:variance-decomposition}

\end{wrapfigure}

We now analyze the relationship between task capability and alignment. Models that perform tasks well may also trivially identify them well, in which case our evaluation setting would add little beyond existing capability evaluations. We test this by exploiting the capability profiles of our three settings (Section~\ref{subsec:setups}). We use Shopping as a capability-saturated control, while we use GDPVal and Terminal-Bench to study alignment under varied capability. CommitRank is our fine-grained metric here.

\paragraph{Task alignment is a separable axis from capability.} 
The Shopping control demonstrates separability directly: all 14 models saturate the underlying task (NDCG spread of $0.07$; Figure~\ref{fig:zero-shot}, Appendix~\ref{section:full-metrics}), while their alignment varies almost threefold, with CommitRank ranging from $0.28$ to $0.81$. Models all capable of performing the task differ widely at identifying it.

The unsaturated settings show that this separability persists even when capability varies. Decomposing the between-model CommitRank variance per task (Figure~\ref{fig:variance-decomposition}), we find that capability explains a meaningful share of CommitRank variance ($43\%$ on GDPVal, $p = 0.011$; $50\%$ on Terminal-Bench, $p = 0.005$). However, a substantial $39\%$ of variance on both tasks is explained by a capability-orthogonal component (bootstrap $p_{\le 0} = 0.017$ on Terminal-Bench, $0.055$ on GDPVal), suggesting that task alignment has its own axis of model behavior and is not just a capability byproduct, requiring separate measurement.

\paragraph{The alignment axis is domain-general.}
\label{paragraph:alignment-domain-general}
Having established that alignment is separable from capability, we now test if the alignment axis is general. We find that models which identify the correct task in one domain tend to do so in other domains .
Concordance of per-task CommitRank rankings across the $14$ models is Kendall's $W = 0.62$ (permutation $p = 0.006$; robust to leave-one-family-out within $[0.43, 0.68]$). To ensure this agreement is not just the capability-shared component of task alignment, we recompute $W$ on capability-residualized ranks on every task. The residual $W = 0.58$ (permutation $p = 0.017$; leave-one-family-out $[0.49, 0.64]$). We thus find that task alignment is a stable axis of model behavior that our framework measures across domains.

\subsection{Human Evaluation}

\begin{figure}[!htbp]
  \centering
  \begin{subfigure}[b]{0.49\linewidth}
    \centering
    \includegraphics[width=\linewidth]{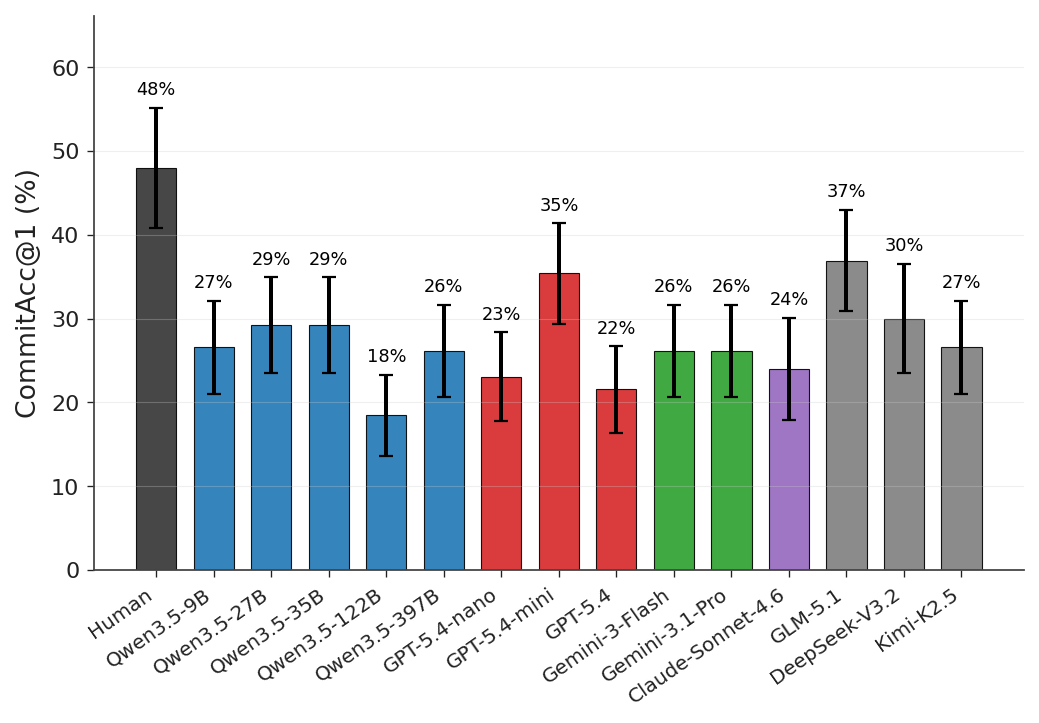}
    \caption{Per-source CommitAcc@1 on the matched Shopping concrete-tier user-study items ($5$ products). 
    Error bars are SEM.}
    \label{fig:human-vs-models-at1}
  \end{subfigure}\hfill
  \begin{subfigure}[b]{0.49\linewidth}
    \centering
    \includegraphics[width=\linewidth]{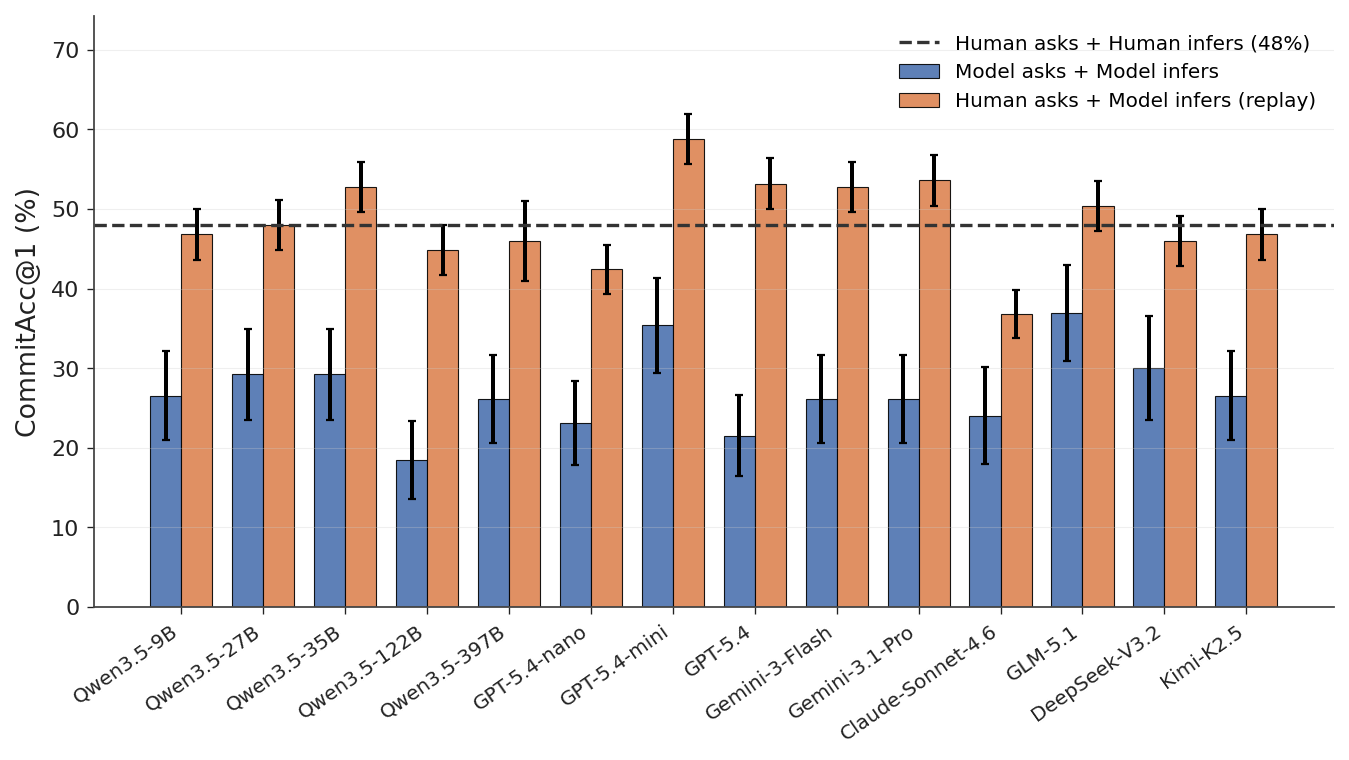}
    \caption{Inference-replay CommitAcc@1. Models with the human's transcript reach a $49.5\%$ median Acc@1; models' inferences on their own transcripts reach only around $25\%$
    }
    \label{fig:replay-comparison-at1}
  \end{subfigure}
  \caption{Human evaluation: (a) humans outperform all models at task recovery; (b) the advantage is explained by communication, not inference.}
  \label{fig:human-eval}
\end{figure}

\paragraph{Humans outperform all LLMs at task recovery.}
\label{subsec:human-asks}
To measure human performance at task alignment, we replace the LLM assistant with a human assistant. We use Shopping as our task setting, since it requires less specialized expertise; and we evaluate in the concrete user tier. We find that humans determine the correct task $48.0 \pm 7.1\%$ of the time (CommitAcc@1) on a smaller dataset of 5 products (Figure~\ref{fig:human-vs-models-at1}), beating the strongest LLM (GLM-5.1 at $36.9 \pm 6.0\%$) and far exceeding median model performance at $\sim25\%$. 
See Tables~ \ref{tab:per-item-human-vs-models-at1-concrete} and ~\ref{tab:per-item-human-vs-models-concrete} in Appendix~\ref{sec:app-user-study} for full details.

\paragraph{Communication ability, not inference, explains the human advantage.}

Given that humans have higher CommitAcc@1 than models (Section~\ref{subsec:human-asks}), we investigate why. We perform an ablation where we have LLMs guess the task conditioned on the human assistant transcripts.
 We find that median model CommitAcc@1 almost doubles to $49.5\%$ from $25\%$, with 8 models exceeding the human assistant + human guess performance which had previously beat all models. $12$ of $14$ models lift their CommitRank (median gain $+0.14$).
The significant lift from using human instead LLM assistant conversations indicates that the human advantage is in communication: \textit{humans are more effective at interacting to resolve and extract information under ambiguity.} Inference ability is not the bottleneck.

\paragraph{Human behavior in specification is captured by our user simulator.}

To validate our user simulator, we also test replacing our LLM simulator with real human users, and we replay our evaluation on three assistant LLMs to compare user conditions. 
Pooled across assistant LLMs, we find that CommitReward (0.98 human, 0.95 simulated), CommitAcc@1 (40.9\% human, 42.0\% simulated), and $T^*$ (13.24 human, 12.45 simulated) are all comparable and within 1 SEM. This study was done post hoc after finalizing our simulator.
Full study details and results are in Appendix~\ref{sec:app-user-study}, and an ablation of our user simulator is in Appendix~\ref{subsec:intent-evolver-ablation}.

\section{Fine-tuning models to improve task alignment}

\begin{wraptable}{r}{0.52\linewidth}
  \vspace{-8pt}
  \centering
  \footnotesize
  \setlength{\tabcolsep}{3.5pt}
  \renewcommand{\arraystretch}{1.25}
  \begin{tabular}{l rrr}
    \toprule
    & base & SFT & RL \\
    \midrule
    \multicolumn{4}{@{}l}{\emph{Task recovery}} \\
    CommitReward           & $0.24_{\pm{.03}}$ & $0.61_{\pm{.03}}$ & $\mathbf{0.62_{\pm{.03}}}$ \\
    CommitAcc@1 (\%)       & $15.2_{\pm{1.4}}$ & $19.6_{\pm{2.3}}$ & $\mathbf{23.6_{\pm{2.4}}}$ \\
    $\mathbb{E}[T^\star]$  & $\mathbf{8.9_{\pm{.1}}}$ & $7.2_{\pm{.2}}$ & $7.9_{\pm{.2}}$ \\
    \midrule
    \multicolumn{4}{@{}l}{\emph{Uncertainty resolution}} \\
    TotalIG                & $\mathbf{0.29_{\pm{.02}}}$ & $0.04_{\pm{.03}}$ & $0.15_{\pm{.02}}$ \\
    Conf.\ corr.\ (\%)     & $11.0$ & $9.0$ & $\mathbf{16.9}$ \\
    Lucky corr.\ (\%)      & $4.0$ & $\mathbf{10.6}$ & $6.7$ \\
    Conf.\ wrong (\%)      & $\mathbf{53.0}$ & $30.9$ & $18.8$ \\
    Hon.\ unc.\ (\%)       & $32.0$ & $49.5$ & $\mathbf{57.5}$ \\
    \bottomrule
  \end{tabular}
    \caption{Training head-to-head for Qwen3.5-9B. \textbf{Top}: outcome metrics (evaluation CommitReward). \textbf{Bottom}: TotalIG and quadrant \%.}
  \label{tab:training}
  \vspace{-8pt}
\end{wraptable}

\label{section:model-training}
We fine-tune Qwen3.5-9B using \textbf{SFT} and \textbf{RL} on Shopping with CommitReward, and evaluate on a matched $n \approx 300$ held-out abstract-tier subset against the base model. Both methods lift CommitReward roughly $2.5\times$ over base, with RL and SFT adding $8.4$ and $4.4$ points on CommitAcc@1  over base (Table~\ref{tab:training},
\emph{Task recovery}). The belief signature also points to training improvements (Table~\ref{tab:training}, \emph Uncertainty resolution). Training shifts mass out of \emph{confidently wrong} ($53.0 \to 30.9\%$ under SFT and $53.0 \to 18.8\%$ under RL, a $34.2$-point drop) and into \emph{honestly uncertain} ($32.0 \to 49.5\%$ under SFT and $32.0 \to 57.5\%$ under RL, $+25.5$ points), while raising at least one correct quadrant. The trained model improves task identification and shifts its certainty to be more honest and calibrated. CommitAcc@1 on human assistant subset for both SFT ($22.5\%$) and RL ($27.5\%$) fall below human performance. Full training details are in Appendix~\ref{sec:app-model-training}.

\section{Related Work}

\paragraph{Task Alignment.} Autonomous agents must determine which task
should be executed, not just execute it. We use \emph{task alignment} to
refer to establishing a shared understanding between user and agent about
the intended task, rooted in the grounding framework of
\citet{clark1991grounding}. Recent work has revisited this idea for autonomous
agents: \citet{wang2025position} positioned it as central to human-agent software engineering. Yet task alignment has not been formalized or measured as its own capability. Meanwhile, a parallel and rich line of work studies personalized alignment and inferring user-specific preferences through multi-turn dialogue \citep{wu2024aligningllmsindividualpreferences,
li2025prefpalette, handa2024bayesianpreferenceelicitationlanguage, qian2024tell, luo2025clarifymtbenchbenchmarkingimprovingmultiturn, montazeralghaem2025askingclarifyingquestionspreference, kim2026discoverllmexecutingintentsdiscovering, zhao2025llmsrecognizepreferencesevaluating}. We apply this line of work from user modeling to formalize and measure task alignment as a new evaluation paradigm, isolated from downstream task execution ability.

\paragraph{Uncertainty Resolution.} A growing body of work studies
clarification under ambiguity: when, what, and how to ask
\citep{zhang2025clarify, li2025questbenchllmsaskright}, recovering implicit
intent in agentic settings \citep{qian2024tell,
vijayvargiya2026ambigsweinteractiveagentsovercome}, and realistic multi-turn
clarification \citep{luo2025clarifymtbenchbenchmarkingimprovingmultiturn}.
Another line improves question-asking via post-training---future-turn
preference labeling \citep{zhang2025modelingfutureconversationturns},
contrastive self-training \citep{chen2025learning}, self-improvement with
simulated preferences \citep{andukuri2024star}, and RL for proactive
information gathering \citep{huang2025teaching}---while multi-turn agent RL
frameworks highlight both the promise and instability of trajectory-level
optimization \citep{wang2025ragen}. We focus instead on \emph{measuring}
uncertainty resolution itself: our metrics track whether interaction improves
an agent's belief over latent task specifications, and how SFT and RL affect
this process.

\section{Conclusion}
We have introduced a domain-general framework for converting fully-specified tasks into task alignment evaluations, formalized as a POMDP over latent tasks. Across shopping, coding, and professional work settings, we find that current models struggle to align with underspecified users; that task alignment is a separable, domain-general axis of model behavior rather than a byproduct of task capability; and that humans outperform all models at identifying users' tasks due to interaction ability. Post-training with SFT and RL improves task recovery and shifts model certainty to be more calibrated, yet a substantial gap in communicative behavior remains. As models improve in capabilities to perform more tasks, our results suggest that task alignment---detecting ambiguity, eliciting information, and knowing when to act---is becoming a primary bottleneck for useful agency, and our framework offers a way to measure and train for it on any specifiable task.

\section*{Limitations}
In this paper, we study LLMs' capability to align on users' tasks through interacting with users. Our study has the following limitations.
\textbf{(1) Training scope.} Due to computational constraints, our SFT and RL experiments are limited to a single open-source backbone (Qwen3.5-9B-Instruct) and a single task setting (Shopping). Whether the gains we observe transfer to larger backbones, to closed-source models, or to the more capability-bound settings (GDPVal, Terminal-Bench) remains open. We expect the qualitative finding that interaction efficiency is the primary axis recovered by post-training to generalize, but the magnitude of the gap to humans may differ substantially in other regimes.

\textbf{(2) User simulation.} Our evaluation relies on an LLM-based user simulator with privileged access to $x^*$ and a stateful intent evolver $E$. Although our matched human-vs-simulator study (Table~\ref{tab:user-sim-val}) shows that the simulator produces broadly similar assistant-side behavior in specification, the simulator cannot capture the full distribution of real user behaviors---genuine task uncertainty, preference drift, contradiction, distraction, frustration with the assistant, or selective disclosure for privacy reasons. In our human-vs-simulator study, humans specify tasks that they have completed already, which conceptually maps roughly to our concrete intent tier, where the task is already known: however, it is not possible to capture abstract intent in humans, limiting the scope of our validation. The intent evolver ablation (Table~\ref{tab:intent-evolver-ablation}) further shows that the choice of evolver materially affects measured performance, meaning our absolute numbers are tied to a particular simulator design rather than a canonical user model.

\bibliography{custom}
\bibliographystyle{colm2026_conference}

\appendix

\section{User study protocol and details}
\label{sec:app-user-study}
\begin{figure}[!htbp]
  \centering
  \includegraphics[width=0.78\linewidth]{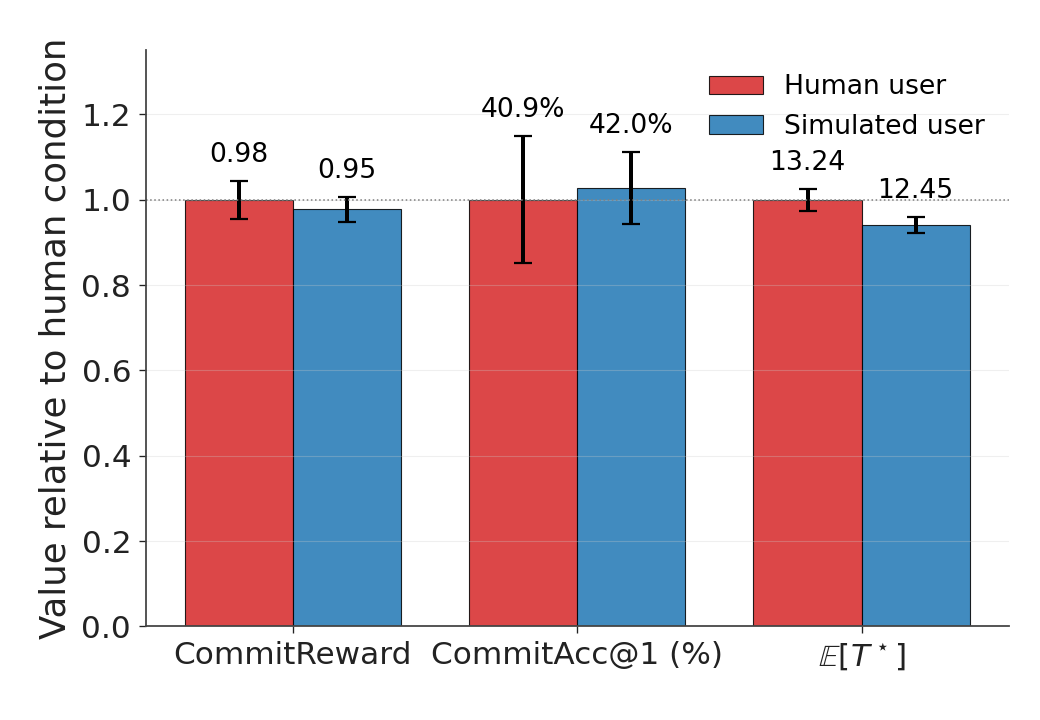}
  \caption{Human vs.\ simulated user, pooled across the three assistants (\texttt{gpt-5.4}, \texttt{qwen3.5-9b}, \texttt{qwen3.5-397b-a17b}) by equal-weight mean of per-assistant means. Each metric (CommitReward, CommitAcc@1, $\mathbb{E}[T^\star]$) is normalized to the human-condition value. Human bars all sit at 1.0; the dotted reference line marks parity. Error bars are pooled SEM in the same normalized scale.}
  \label{fig:sim-reward}
\end{figure}

We launch two human user studies on \href{https://www.prolific.com}{Prolific} to collect (1) human-as-assistant (Table~\ref{tab:per-item-human-vs-models-concrete}; Figure~\ref{fig:human-vs-models-at1}), and (2) human-as-user baselines. The goal of human-as-assistant is to compare model and human performance, while the goal of human-as-user is to validate our user simulator (Table~\ref{tab:user-sim-val}; Figure~\ref{fig:sim-reward}). We select Shopping as the task setting since it is accessible for humans to participate, and also easily scalable.

Human participants in our studies were filtered with three pre-screening criteria: (1) fluent English, (2) location to the United States or United Kingdom, and (3) Prolific's verification for completing AI studies. Each participant completed a single task. Compensation was an effective rate of approximately \$$12$/hour, with the per-session rate set against the pilot median completion time. Both studies shared the same recruitment pool, but no participant was allowed to appear in both studies. To guarantee the privacy of participants, we instructed participants to not provide any personally identifiable information (PII) and provided a disclaimer of the goals and how the transcripts would be used.

\subsection{Human-as-assistant}
\label{subsec:human-as-assistant}
For the human-as-assistant experiments, we launch our experiments in Section~\ref{subsec:setups} but have humans interact with our LLM user simulators. The human interacts with the LLM user simulator initialized with concrete intent for up to 15 turns, then ranks the 15 candidate items. We run 50 sessions across 5 products with 50 participants from Prolific. Tables~\ref{tab:per-item-human-vs-models-at1-concrete} and ~\ref{tab:per-item-human-vs-models-concrete} contain the per-item user simulator breakdown for all the assistants (both model and human).

\begin{table*}[!htbp]
\scriptsize
\centering
\caption{Per-item mean CommitAcc@1 (\%, $\pm$ standard error) on the Concrete intent tier for the human-asks user study vs.\ each assistant model. Each model cell aggregates $\approx 13$ rollouts; each human cell aggregates the corresponding user-study sessions on that item.}
\label{tab:per-item-human-vs-models-at1-concrete}
\begin{tabular}{l c c c c c c}
\toprule
\textbf{Source} & \textbf{antacid} & \textbf{wellies} & \textbf{travel mug} & \textbf{dog treats} & \textbf{toiletry bag} & \textbf{Overall} \\
\midrule
\textbf{HUMAN} & $66.7 \pm 14.2$ & $50.0 \pm 15.1$ & $18.2 \pm 12.2$ & $66.7 \pm 16.7$ & $33.3 \pm 21.1$ & $48.0 \pm 7.1$ \\
\midrule
Qwen3.5-9B & $7.7 \pm 7.7$ & $15.4 \pm 10.4$ & $0.0 \pm 0.0$ & $92.3 \pm 7.7$ & $16.7 \pm 11.2$ & $26.6 \pm 5.6$ \\
Qwen3.5-27B & $15.4 \pm 10.4$ & $7.7 \pm 7.7$ & $0.0 \pm 0.0$ & $84.6 \pm 10.4$ & $38.5 \pm 14.0$ & $29.2 \pm 5.7$ \\
Qwen3.5-35B & $7.7 \pm 7.7$ & $15.4 \pm 10.4$ & $0.0 \pm 0.0$ & $\mathbf{100.0 \pm 0.0}$ & $23.1 \pm 12.2$ & $29.2 \pm 5.7$ \\
Qwen3.5-122B & $0.0 \pm 0.0$ & $0.0 \pm 0.0$ & $0.0 \pm 0.0$ & $92.3 \pm 7.7$ & $0.0 \pm 0.0$ & $18.5 \pm 4.8$ \\
Qwen3.5-397B & $7.7 \pm 7.7$ & $15.4 \pm 10.4$ & $0.0 \pm 0.0$ & $84.6 \pm 10.4$ & $23.1 \pm 12.2$ & $26.2 \pm 5.5$ \\
GPT-5.4-nano & $7.7 \pm 7.7$ & $7.7 \pm 7.7$ & $\mathbf{7.7 \pm 7.7}$ & $61.5 \pm 14.0$ & $30.8 \pm 13.3$ & $23.1 \pm 5.3$ \\
GPT-5.4-mini & $15.4 \pm 10.4$ & $7.7 \pm 7.7$ & $0.0 \pm 0.0$ & $84.6 \pm 10.4$ & $\mathbf{69.2 \pm 13.3}$ & $35.4 \pm 6.0$ \\
GPT-5.4 & $0.0 \pm 0.0$ & $0.0 \pm 0.0$ & $0.0 \pm 0.0$ & $100.0 \pm 0.0$ & $7.7 \pm 7.7$ & $21.5 \pm 5.1$ \\
Gemini-3-Flash & $7.7 \pm 7.7$ & $15.4 \pm 10.4$ & $0.0 \pm 0.0$ & $100.0 \pm 0.0$ & $7.7 \pm 7.7$ & $26.2 \pm 5.5$ \\
Gemini-3.1-Pro & $0.0 \pm 0.0$ & $\mathbf{23.1 \pm 12.2}$ & $0.0 \pm 0.0$ & $100.0 \pm 0.0$ & $7.7 \pm 7.7$ & $26.2 \pm 5.5$ \\
Claude-Sonnet-4.6 & $0.0 \pm 0.0$ & $0.0 \pm 0.0$ & $0.0 \pm 0.0$ & $100.0 \pm 0.0$ & $20.0 \pm 13.3$ & $24.0 \pm 6.1$ \\
GLM-5.1 & $\mathbf{38.5 \pm 14.0}$ & $7.7 \pm 7.7$ & $7.7 \pm 7.7$ & $100.0 \pm 0.0$ & $30.8 \pm 13.3$ & $\mathbf{36.9 \pm 6.0}$ \\
DeepSeek-V3.2 & $20.0 \pm 13.3$ & $0.0 \pm 0.0$ & $0.0 \pm 0.0$ & $70.0 \pm 15.3$ & $60.0 \pm 16.3$ & $30.0 \pm 6.5$ \\
Kimi-K2.5 & $0.0 \pm 0.0$ & $0.0 \pm 0.0$ & $0.0 \pm 0.0$ & $100.0 \pm 0.0$ & $38.5 \pm 14.0$ & $26.6 \pm 5.6$ \\
\bottomrule
\end{tabular}
\end{table*}
  
\begin{table*}[!htbp]
\scriptsize
\centering
\caption{Per-item mean reward ($\pm$ standard error) on the Concrete intent tier for the human-asks user study vs.\ each assistant model. Each model cell aggregates $\approx 13$ rollouts; each human cell aggregates the corresponding user-study sessions on that item.}
\label{tab:per-item-human-vs-models-concrete}
\begin{tabular}{l c c c c c c}
\toprule
\textbf{Source} & \textbf{antacid} & \textbf{wellies} & \textbf{travel mug} & \textbf{dog treats} & \textbf{toiletry bag} & \textbf{Overall} \\
\midrule
\textbf{HUMAN} & $0.733 \pm 0.083$ & $0.641 \pm 0.066$ & $0.563 \pm 0.094$ & $0.675 \pm 0.087$ & $0.560 \pm 0.131$ & $0.642 \pm 0.039$ \\
\midrule
Qwen3.5-9B & $1.025 \pm 0.120$ & $0.618 \pm 0.181$ & $0.166 \pm 0.100$ & $1.386 \pm 0.012$ & $0.568 \pm 0.181$ & $0.758 \pm 0.078$ \\
Qwen3.5-27B & $0.847 \pm 0.134$ & $0.970 \pm 0.120$ & $0.573 \pm 0.137$ & $1.322 \pm 0.042$ & $1.241 \pm 0.077$ & $0.990 \pm 0.058$ \\
Qwen3.5-35B & $1.003 \pm 0.104$ & $0.742 \pm 0.160$ & $0.495 \pm 0.132$ & $1.380 \pm 0.011$ & $1.199 \pm 0.048$ & $0.964 \pm 0.061$ \\
Qwen3.5-122B & $0.545 \pm 0.143$ & $0.232 \pm 0.097$ & $0.384 \pm 0.129$ & $1.422 \pm 0.020$ & $0.933 \pm 0.138$ & $0.703 \pm 0.073$ \\
Qwen3.5-397B & $0.849 \pm 0.121$ & $0.980 \pm 0.133$ & $0.442 \pm 0.115$ & $1.301 \pm 0.101$ & $1.232 \pm 0.086$ & $0.961 \pm 0.062$ \\
GPT-5.4-nano & $0.321 \pm 0.126$ & $0.816 \pm 0.151$ & $0.518 \pm 0.112$ & $1.015 \pm 0.176$ & $1.117 \pm 0.116$ & $0.757 \pm 0.071$ \\
GPT-5.4-mini & $0.954 \pm 0.082$ & $0.769 \pm 0.126$ & $0.778 \pm 0.093$ & $1.375 \pm 0.081$ & $1.334 \pm 0.114$ & $1.042 \pm 0.055$ \\
GPT-5.4 & $0.973 \pm 0.042$ & $0.640 \pm 0.091$ & $0.760 \pm 0.075$ & $\mathbf{1.477 \pm 0.002}$ & $\mathbf{1.366 \pm 0.008}$ & $1.043 \pm 0.048$ \\
Gemini-3-Flash & $0.764 \pm 0.066$ & $\mathbf{1.132 \pm 0.118}$ & $0.901 \pm 0.069$ & $1.446 \pm 0.005$ & $1.246 \pm 0.046$ & $\mathbf{1.098 \pm 0.043}$ \\
Gemini-3.1-Pro & $\mathbf{1.115 \pm 0.067}$ & $0.671 \pm 0.161$ & $\mathbf{1.088 \pm 0.019}$ & $1.431 \pm 0.004$ & $1.085 \pm 0.112$ & $1.078 \pm 0.050$ \\
Claude-Sonnet-4.6 & $0.609 \pm 0.122$ & $0.706 \pm 0.155$ & $0.597 \pm 0.123$ & $1.430 \pm 0.003$ & $1.284 \pm 0.029$ & $0.925 \pm 0.068$ \\
GLM-5.1 & $1.028 \pm 0.119$ & $0.497 \pm 0.103$ & $0.819 \pm 0.114$ & $1.428 \pm 0.003$ & $1.355 \pm 0.018$ & $1.025 \pm 0.057$ \\
DeepSeek-V3.2 & $0.668 \pm 0.191$ & $0.813 \pm 0.144$ & $0.015 \pm 0.022$ & $1.120 \pm 0.165$ & $1.360 \pm 0.021$ & $0.795 \pm 0.086$ \\
Kimi-K2.5 & $0.413 \pm 0.099$ & $0.567 \pm 0.102$ & $0.719 \pm 0.118$ & $1.455 \pm 0.004$ & $1.112 \pm 0.151$ & $0.844 \pm 0.067$ \\
\bottomrule
\end{tabular}
\end{table*}

\subsection{Human-as-user}
\label{subsec:human-as-user}
For the human-as-user experiments, we run our experiments in Section~\ref{subsec:setups} with human users, and then backfill LLM user simulators for comparison using the process described in Section~\ref{subsec:task-settings}. Between the two user conditions, we compare performance of three LLM assistants (\texttt{gpt-5.4}, \texttt{qwen3.5-9b}, \texttt{qwen3.5-397b-a17b}) at aligning with both user groups. Table~\ref{tab:user-sim-val} contains the comparisons across the three assistant models. For each assistant, we assign a pool of $n=25$ human users on Prolific. We again use Shopping as our task setting for the same reasons as in Section~\ref{subsec:human-asks}. The ground-truth task $x^*$ is determined by the human user who provides an Amazon.com item from their purchase history. 

Given an item $x^*$ from the human user, we create the corresponding LLM user simulator by following the procedure in Section~\ref{subsec:task-settings} to initialize the intent state and generate the candidate set. Each human user gives a task $x^*$, and we create a simulator for each $x^*$ to create the LLM user set, divided into pools of $n=25$ based on which assistant the human user interacted with. We then run our task alignment evaluation with the LLM user simulators on the three LLM assistants (\texttt{gpt-5.4}, \texttt{qwen3.5-9b}, \texttt{qwen3.5-397b-a17b}).

\begin{table}[!htbp]
\centering
\caption{Matched simulator-validation comparison: same LLM as the asker, two user conditions. Real humans came with a specific product
in mind (similar to concrete intent), so the matched sim condition is also treated as concrete-tier.}
\label{tab:user-sim-val}
\setlength{\tabcolsep}{3.8pt}
\begin{tabular}{ll rrrr}
  \toprule
  Assistant & User & CommitReward & CommitAcc@1 (\%) & $\mathbb{E}[T^\star]$ \\
  \midrule
  \texttt{gpt-5.4} & human & $1.15 \pm 0.04$ & $50.00 \pm 11.47$ & $11.90 \pm 1.02$  \\
  \texttt{gpt-5.4} & sim & $0.98 \pm 0.05$ & $36.51 \pm 6.11$ & $10.24 \pm 0.70$\\
  \midrule
  \texttt{qwen3.5-9b} & human & $0.80 \pm 0.10$ & $29.17 \pm 9.48$ & $14.00 \pm 0.00$  \\
  \texttt{qwen3.5-9b} & sim & $0.89 \pm 0.05$ & $41.67 \pm 5.85$ & $13.13 \pm 0.28$ \\
  \midrule
  \texttt{qwen3.5-397b} & human & $0.98 \pm 0.08$ & $43.48 \pm 10.57$ & $13.83 \pm 0.17$  \\
  \texttt{qwen3.5-397b} & sim & $0.99 \pm 0.04$ & $47.83 \pm 6.06$ & $14.00 \pm 0.00$ \\
  \bottomrule
\end{tabular}
\end{table}

\subsection{Intent evolver ablation}
\label{subsec:intent-evolver-ablation}

\paragraph{Intent evolver is essential for simulating human behavior.}
\label{subsec:intent-evolver-ablation}
We run an ablation removing the intent evolver from our user simulator. Table~\ref{tab:intent-evolver-ablation} shows that removing the intent evolver raises CommitReward across all three assistants (GPT-5.4: $0.98\!\to\!1.22$; Qwen3.5-9B: $0.89\!\to\!1.17$; Qwen3.5-397B: $0.99\!\to\!1.19$) and shortens $\mathbb{E}[T^\star]$ (GPT-5.4: $10.24\!\to\!8.10$), indicating that without gradual intent evolution the simulator becomes too transparent and models can commit earlier and more accurately.

The matched comparison with human user behavior in specifying tasks depends on the simulator's behavior being  gradual: the simulator must not fully know the intent and discover it incrementally. The intent evolver controls which aspects of the underlying intent are exposed at each turn if it is triggered in the conversation, which is likely what results in our simulator being plausibly human in simulating specification.

 \begin{table}[!htbp]
    \centering
    \small
    \caption{Intent evolver ablation on Shopping for concrete tier. Both conditions use the same (concrete-tier) initial intent; the only difference is whether the sim's intent is re-evolved each turn from the conversation. Disabling the evolver freezes the intent at the fully specified concrete seed.}
\label{tab:intent-evolver-ablation}
  \begin{tabular}{ll rrr}
    \toprule
    Assistant & Condition & CommitReward & $\mathbb{E}[T^\star]$ \\
    \midrule
    \texttt{gpt-5.4} & with-evolver & $1.00 \pm 0.05$ & $10.24 \pm 0.70$ \\
    \texttt{gpt-5.4} & no-evolver & $1.22 \pm 0.05$ & $8.10 \pm 1.35$ \\
    \midrule
    \texttt{qwen3.5-9b} & with-evolver & $0.92 \pm 0.05$ & $13.13 \pm 0.28$ \\
    \texttt{qwen3.5-9b} & no-evolver & $1.17 \pm 0.06$ & $11.96 \pm 0.70$ \\
    \midrule
    \texttt{qwen3.5-397b-a17b} & with-evolver & $1.00 \pm 0.04$ & $14.00 \pm 0.00$ \\
    \texttt{qwen3.5-397b-a17b} & no-evolver & $1.19 \pm 0.02$ & $13.74 \pm 0.26$ \\
    \bottomrule
  \end{tabular}
\end{table}

\paragraph{Conversation alignment with the ground-truth specification.}
We track how much of the ground-truth specification has been surfaced
by turn~$t$ via \emph{spec-token coverage}, the fraction of unique
content tokens in the reference specification that appear in the
cumulative user-revealed text from turns $1$ to $t$. 

\begin{figure}[!htbp]
  \centering
  \includegraphics[width=0.5\linewidth]{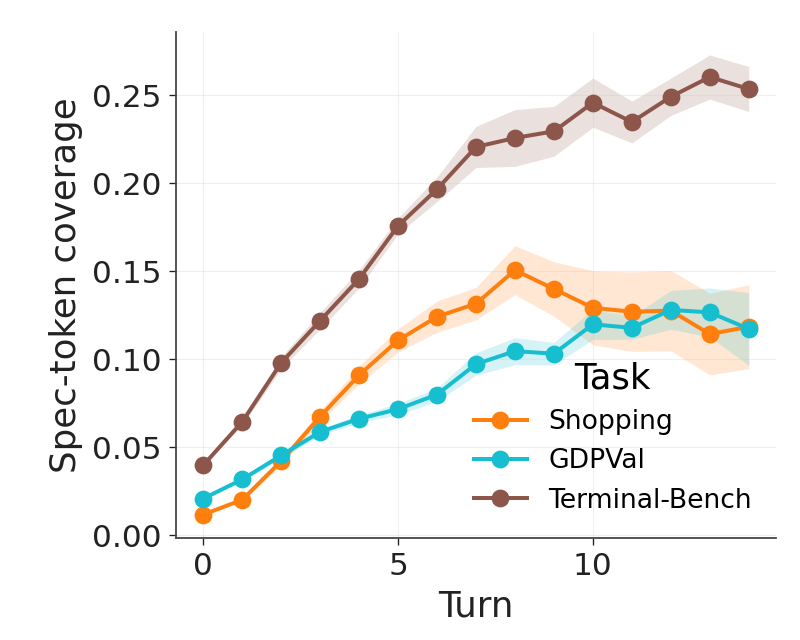}
  \caption{Per-turn spec-token coverage = $|U_t \cap S|/|S|$, where $U_t$ is the set of unique content tokens in the cumulative user text up to turn $t$ and $S$ is the set of unique content tokens in the ground-truth specification. Averaged across the 14 models at the abstract tier; shadings are $\pm$SEM across models.}
  \label{fig:bleu-over-turns}
\end{figure}

Concretely, $\text{coverage}_t = |U_t \cap S| / |S|$, where $U_t$ is the unique content-token set of the cumulative user text and $S$ is the unique content-token set of the task specification. Figure~\ref{fig:bleu-over-turns} averages this across all 14 models at the abstract tier. All three tasks rise monotonically through conversation despite having an initially abstract intent state, suggesting that the intent evolver is guiding the sequence of intent states toward $x^*$ as expected.

\newpage
\section{Task setting creation}
\label{section:task-setting-creation}

Task setting creation turns each fully specified benchmark task $x^\star$ into a size-$15$ candidate set $\mathcal{C}$ plus abstract/moderate/concrete intent seeds for the user simulator. The procedure has three conceptual stages, instantiated with the same shape for Shopping, GDPVal, and Terminal-Bench. The difference is whether the values are drawn from a existing catalog (Shopping) or synthesized by prompted LLM calls (GDPVal, Terminal-Bench, prompts in Appendix~\ref{sec:app-prompts}) where catalogs of tasks are harder to curate.

\paragraph{Stage 1: Identify perturbation axes.} Enumerate the dimensions along which a user could plausibly want a variant of $x^\star$. For \textbf{Shopping}, the axes are the features behind the underlying search query (e.g.\ character/theme, brand, target audience for ``inflatable costume'').  For \textbf{GDPVal} and \textbf{Terminal-Bench}, one LLM call reads $x^\star$ and its rubric (or test code) and returns the binding axes together with $14$ delta-sets: the caffe-cifar-10 example yields ten axes ranging from Caffe fork to test-accuracy floor.

\paragraph{Stage 2: Populate candidates by sampling fresh values along a subset of the axes.} For \textbf{Shopping}, the first $14$ Amazon search results for the query naturally sample real products along those axes. For \textbf{GDPVal} and \textbf{Terminal-Bench}, an LLM rewrites $x^\star$ per candidate with the new sampled values substituted in place.

\paragraph{Stage 3: Verify internal coherence.} For \textbf{Shopping}, real listings are naturally self-consistent, so no repair is needed. For \textbf{GDPVal} and \textbf{Terminal-Bench}, an audit pass, along with human manual review, rewrites any candidate whose perturbation introduced internal contradictions or references back to $x^\star$. 

User intent seeds (abstract/moderate/concrete) are then generated by a separate prompt over $x^\star$ that emits progressively less specified initial user messages (\texttt{ABSTRACTIONS\_TEMPLATE}).

\paragraph{Candidate set validation.} A candidate set that leaks $x^\star$ through typical patterns or canonical defaults would let a ranker identify the correct task without any signal, trivializing the interaction. We check for this by running the ranking task over each row with an empty conversation and recording where $x^\star$ ends up (Table~\ref{tab:null-intent-validation}). On benchmarks $x^\star$ sits at or above the random-baseline mean rank of $8.0$ across $15$ candidates ($8.71$ on GDPVal, $9.07$ on Terminal-Bench), top-$1$ rate with empty conversation is around the random $6.7\%$ rate ($3.2\%$ and $6.9\%$), and only $3.2\%$/$6.9\%$ of rows land $x^\star$ at rank $\leq 2$.

\paragraph{Example.} Table~\ref{tab:candidate-generation} shows one worked example per benchmark --- inflatable costumes for Shopping, a real-estate broker Letter-of-Intent for GDPVal, and the caffe-cifar-10 environment-setup task for Terminal-Bench, where each perturbed slot is bolded relative to $x^\star$.

\begin{table}[!htbp]
  \centering
  \small
  \caption{Empty conversation probe on the accepted candidate sets: mean/median rank of $x^\star$ under an empty-conversation ranker, top-$1$ leakage rate, and mean normalized rank $\hat r_{\text{null}}$. Compared against the random-shuffle baseline for $|\mathcal{C}|=15$ candidates.}
  \label{tab:null-intent-validation}
  \setlength{\tabcolsep}{5pt}
  \begin{tabular}{l r rr rr r}
    \toprule
    & $n$ & mean rank & median & top-$1$ (\%) & $\hat r_{\text{null}}$ & rank $\leq 2$ (\%) \\
    \midrule
    GDPVal          & $95$ & $8.71$ & $8$ & $3.2$ & $0.429$ & $3.2$ \\
    Terminal-Bench  & $87$ & $9.07$ & $9$ & $6.9$ & $0.406$ & $6.9$ \\
    \bottomrule
  \end{tabular}
\end{table}

\begin{table*}[!htbp]
\tiny
\centering
\caption{Full candidate-set examples per benchmark. Values that differ from $x^\star$ along a perturbation dimension are \textbf{bolded}.}
\label{tab:candidate-generation}
\begin{tabular}{@{}p{2.0cm} p{13.0cm}@{}}
\toprule
\multicolumn{2}{@{}l@{}}{\textbf{Shopping} \emph{(query: ``inflatable costume''; 15 real Amazon search results)}} \\
\midrule
\textbf{$x^\star$} & \textbf{Bodysocks} \textbf{Red Devil} Inflatable Costume for \textbf{Adults} (One Size) \\
\textbf{Axes (3)} & brand; character/theme; target audience. \\
\midrule
Candidate 1 & \textbf{Rubies} Jurassic World: \textbf{T-Rex} Inflatable Costume for \textbf{Adults}, Unisex One Size \\
\midrule
Candidate 2 & \textbf{miwhse} Halloween Inflatable \textbf{Dinosaur} Costume \textbf{Adult} \ldots{} Full Body Dino Costume For Halloween Cosplay Party \\
\midrule
Candidate 3 & \textbf{Boveco} \textbf{Red Love Heart} Inflatable Costume Adult \textbf{Valentines Day} Blow Up Mascot Costume \ldots{} \\
\midrule
Candidate 4 & \textbf{KOOY} Inflatable \textbf{Parrot} Costume \textbf{Kids} \ldots{} Ride On Inflatable Costume For Kids \\
\midrule
Candidate 5 & \textbf{HIYAPATY} Inflatable \textbf{Bumble Bee} Costume \textbf{Men Women} \ldots{} Adult Funny Halloween Insects Outfit \\
\midrule
Candidate 6 & \textbf{Rubies} \textbf{Astronaut} Inflatable Costume for \textbf{Kids}, Unisex One Size \\
\midrule
Candidate 7... & \textbf{LookOurWay} Inflatable \textbf{Tube Man} Costume \ldots{} Wacky Wavy Arm Guy Funny Inflatable Halloween Costume for Adults \\
\midrule

\multicolumn{2}{@{}l@{}}{\textbf{GDPVal} \emph{(Real Estate Broker: draft a Letter of Intent in Word for a commercial-property acquisition)}} \\
\midrule
\textbf{$x^\star$} & \ldots{} client \textbf{Annocium Investors} \ldots{} \textbf{536--41 Fraanklyn Ave, Denver} \ldots{} \textbf{1031 exchange} \ldots{} advertised by \textbf{Bob Crobens of HPTR} for \textbf{\$9{,}000{,}000} @ \textbf{6\%} cap \ldots{} reflect \textbf{6.5\%} cap \ldots{} seller \textbf{Denver Services Bank} \ldots{} \textbf{90-day} feasibility, closing \textbf{+90 days}, extension \textbf{\$20{,}000} \ldots{} initial \textbf{\$100{,}000}, additional \textbf{\$150{,}000} \ldots{} escrow \textbf{First American Title}. \\
\textbf{Axes (12)} & buyer; seller; seller's broker/firm; property address \& acreage; exchange structure; list price \& cap rate; negotiated cap rate; feasibility period; closing period; initial deposit; additional deposit; extension-option deposit; escrow/title company. \\
\midrule
Candidate 1 & client \textbf{Meridian Capital Group} \ldots{} \textbf{1200--08 Larimer St, Denver} \ldots{} \textbf{straight purchase} \ldots{} \textbf{Mark Tillerson, WKRT Realty} \ldots{} \textbf{\$12M} @ \textbf{5.0\%} $\to$ \textbf{5.5\%} \ldots{} seller \textbf{Rocky Mtn.\ Savings Trust} \ldots{} \textbf{60-day} feas., closing \textbf{+60 days}, ext.\ \textbf{\$30K} \ldots{} init \textbf{\$50K}, add'l \textbf{\$100K} \ldots{} escrow \textbf{Chicago Title}. \\
\midrule
Candidate 2 & client \textbf{Vantage Realty Partners} \ldots{} \textbf{720--28 Blake St, Denver} \ldots{} \textbf{721 exchange (UPREIT)} \ldots{} \textbf{Sandra Owens, Pacific Bridge Commercial} \ldots{} \textbf{\$7.5M} @ \textbf{5.5\%} $\to$ \textbf{6.0\%} \ldots{} seller \textbf{Frontier Commercial Realty} \ldots{} \textbf{120-day} feas., closing \textbf{+120 days}, ext.\ \textbf{\$50K} \ldots{} init \textbf{\$150K}, add'l \textbf{\$200K} \ldots{} escrow \textbf{Fidelity National Title}. \\
\midrule
Candidate 3 & client \textbf{Stonecrest Equity Fund} \ldots{} \textbf{1600 Glenarm Pl, Denver} \ldots{} \textbf{1031 exchange} \ldots{} \textbf{Carl Nettles, Apex Commercial Group} \ldots{} \textbf{\$9M} @ \textbf{6\%} $\to$ \textbf{6.5\%} \ldots{} seller \textbf{Colorado National Properties LLC} \ldots{} \textbf{45-day} feas., closing \textbf{+30 days}, ext.\ \textbf{\$10K} \ldots{} init \textbf{\$75K}, add'l \textbf{\$250K} \ldots{} escrow \textbf{Stewart Title}. \\
\midrule
Candidate 4 & client \textbf{Keystone Asset Holdings} \ldots{} \textbf{850--56 Wynkoop St, Denver} \ldots{} \textbf{straight purchase} \ldots{} \textbf{Diane Rafferty, Cascade Realty Advisors} \ldots{} \textbf{\$10.5M} @ \textbf{5.5\%} $\to$ \textbf{6.0\%} \ldots{} seller \textbf{Summit Holdings Corp} \ldots{} \textbf{75-day} feas., closing \textbf{+75 days}, ext.\ \textbf{\$25K} \ldots{} init \textbf{\$200K}, add'l \textbf{\$75K} \ldots{} escrow \textbf{Land Title Guarantee Co.} \\
\midrule
Candidate 5 & client \textbf{Pinnacle Investment Trust} \ldots{} \textbf{720--28 Blake St, Denver} \ldots{} \textbf{1031 exchange} \ldots{} \textbf{Mark Tillerson, WKRT Realty} \ldots{} \textbf{\$12M} @ \textbf{5.0\%} $\to$ \textbf{5.5\%} \ldots{} seller \textbf{Rocky Mtn.\ Savings Trust} \ldots{} \textbf{60-day} feas., closing \textbf{+120 days}, ext.\ \textbf{\$30K} \ldots{} init \textbf{\$50K}, add'l \textbf{\$150K} \ldots{} escrow \textbf{Chicago Title}. \\
\midrule
Candidate 6... & client \textbf{Vantage Realty Partners} \ldots{} \textbf{1200--08 Larimer St, Denver} \ldots{} \textbf{1033 exchange} \ldots{} \textbf{Carl Nettles, Apex Commercial Group} \ldots{} \textbf{\$10.5M} @ \textbf{5.5\%} $\to$ \textbf{6.0\%} \ldots{} seller \textbf{Summit Holdings Corp} \ldots{} \textbf{120-day} feas., closing \textbf{+60 days}, ext.\ \textbf{\$10K} \ldots{} init \textbf{\$150K}, add'l \textbf{\$100K} \ldots{} escrow \textbf{Stewart Title}. \\
\midrule

\multicolumn{2}{@{}l@{}}{\textbf{Terminal-Bench} \emph{(caffe-cifar-10: install BVLC Caffe and train a CIFAR-10 classifier)}} \\
\midrule
\textbf{$x^\star$} & Install the original \textbf{BVLC} Caffe deep learning framework (version \textbf{1.0.0}) and train a CNN to classify \textbf{CIFAR-10} images. Clone Caffe to \textbf{/app/caffe} and build for only \textbf{CPU} execution, training for exactly \textbf{500} iterations. Write the training output to \textbf{/app/caffe/training\_output.txt} and verify that the test accuracy (for \textbf{100} iterations) is no more than \textbf{5\%} less than train and greater than \textbf{45\%}. The model file should be available in the \textbf{examples/cifar10} directory and be named \texttt{cifar10\_quick\_iter\_\{number\_of\_iterations\}.caffemodel}. \\
\textbf{Axes (10)} & caffe fork/version; clone path; build mode; dataset; model architecture; training iterations; training-output filename; train-vs-test accuracy gap; minimum test-accuracy threshold; test-evaluation iteration count. \\
\midrule
Candidate 1 & \ldots{} \textbf{BVLC 1.0.0} \ldots{} classify \textbf{CIFAR-10} images. Clone to \textbf{/app/caffe} and build for only \textbf{CPU}, training for exactly \textbf{1000} iterations. Write to \textbf{/app/caffe/train\_log.txt} and verify test accuracy (for \textbf{100} iterations) is no more than \textbf{5\%} less than train and greater than \textbf{50\%}. \ldots{} \\
\midrule
Candidate 2 & \ldots{} \textbf{BVLC 1.0.0} \ldots{} classify \textbf{CIFAR-10} images. Clone to \textbf{/app/caffe} and build for only \textbf{CPU}, training for exactly \textbf{2000} iterations. Write to \textbf{/app/caffe/output.txt} and verify test accuracy (for \textbf{50} iterations) is no more than \textbf{3\%} less than train and greater than \textbf{55\%}. \ldots{} \\
\midrule
Candidate 3 & \ldots{} \textbf{BVLC 1.0.0} \ldots{} classify \textbf{CIFAR-10} images. Clone to \textbf{/app/caffe} and build for only \textbf{CPU}, training for exactly \textbf{5000} iterations. Write to \textbf{/app/caffe/log.txt} and verify test accuracy (for \textbf{200} iterations) is no more than \textbf{10\%} less than train and greater than \textbf{60\%}. \ldots{} \\
\midrule
Candidate 4 & \ldots{} \textbf{NVIDIA-fork} Caffe (\textbf{0.17}) \ldots{} classify \textbf{MNIST} images. Clone to \textbf{/opt/caffe} and build for \textbf{GPU} execution, training for exactly \textbf{1000} iterations. Write to \textbf{/opt/caffe/mnist\_train.log} and verify test accuracy (for \textbf{100} iterations) is within \textbf{5\%} of train and greater than \textbf{95\%}. \ldots{} Model at \textbf{examples/mnist/lenet\_iter\_\{n\}.caffemodel}. \\
\midrule
Candidate 5 & \ldots{} \textbf{BVLC 1.0.0} \ldots{} classify \textbf{CIFAR-10} images using the \textbf{cifar10\_full} architecture. Clone to \textbf{/workspace/caffe} and build for \textbf{CPU}, training for exactly \textbf{4000} iterations. Write to \textbf{/workspace/caffe/full\_train.log} \ldots{} greater than \textbf{70\%}. \ldots{} \\
\midrule
Candidate 6 & \ldots{} \textbf{BVLC 1.0.0} \ldots{} classify \textbf{CIFAR-10} images. Clone to \textbf{/src/caffe} and build for \textbf{GPU} execution, training for exactly \textbf{500} iterations. Write to \textbf{/src/caffe/training\_output.txt} and verify test accuracy (for \textbf{100} iterations) is no more than \textbf{5\%} less than train and greater than \textbf{45\%}. \ldots{} \\
\midrule
Candidate 7... & \ldots{} \textbf{BVLC 1.0.0} \ldots{} classify \textbf{CIFAR-10} images. Clone to \textbf{/app/caffe} and build for only \textbf{CPU}, training for exactly \textbf{500} iterations. Write to \textbf{/app/caffe/training\_output.txt} \ldots{} no more than \textbf{2\%} less than train and greater than \textbf{55\%} (evaluated for \textbf{500} iterations). \ldots{} \\
\bottomrule
\end{tabular}
\end{table*}

\newpage
\section{Entropy calculation details}
\label{section:entropy-calculation}

We measure the assistant's belief entropy over the candidate set using the ranking task $\rho_t$ over that runs in a separate conversation thread from the interaction with the user, once per turn. There are two implementation details to note: a letter-shuffle that turns the assistant's next-token letter distribution into a distribution over $\mathcal{C}$, and a calibration step that accounts for letter frequency priors and isolates signal. All uncertainty metrics in the paper (per-turn $\tilde H_t$, $\mathrm{TotalIG}$, and the $H$-splits in the four-quadrant taxonomy) follow these steps.

\paragraph{Letter-shuffle.} After each turn we assign $|\mathcal{C}|=15$ candidates to letters $A$--$O$ under a random permutation, then prompt the assistant with the conversation history and the shuffled option list, and instruct the LM to ``\texttt{Answer with ONLY the letter.}''. We read the top-$20$ logprobs at the first response position, pick out $\ell_L$ for each letter $L \in \{A, \ldots, O\}$, and softmax across letters to obtain a per-turn belief $b_t$ over candidates. The point of reshuffling every turn is to break any position bias. The normalized belief entropy is $\tilde H_t = -\sum_c b_{t,c} \log b_{t,c} / \log |\mathcal{C}| \in [0,1]$, and $\mathrm{TotalIG} = \tilde H_0 - \tilde H_{T^\star}$ measures how much of that uncertainty the assistant resolves before commit.

\paragraph{Empty conversation calibration.} To isolate the interaction signal in the candidate ranking $\rho_t$, we rerun the identical $\rho_t$ prompt a second time with the conversation replaced by the placeholder ``\texttt{(no conversation context)}'', keeping the same candidates in the same shuffled positions. This second call gives a baseline logprob $\ell_L^{\text{null}}$ per letter, and we softmax the difference $\tilde\ell_L = \ell_L - \ell_L^{\text{null}}$ to obtain the calibrated belief and its entropy $\tilde H_t^{\text{cal}}$.

\section{Additional results}
\label{section:full-metrics}
\begin{figure}[!htbp]
  \centering
  \small
  \includegraphics[width=0.55\linewidth]{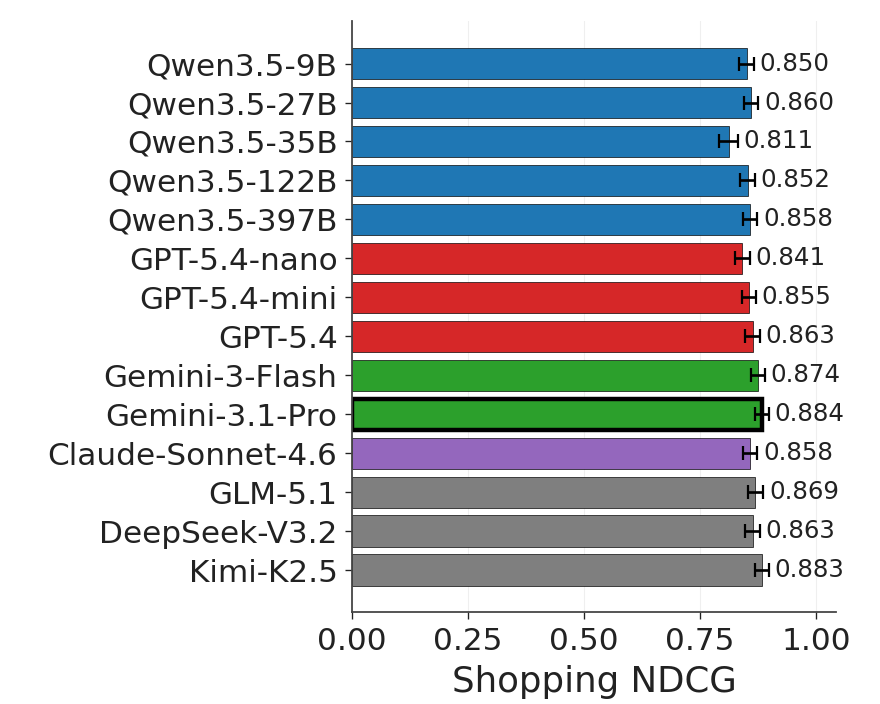}
  \includegraphics[width=0.55\linewidth]{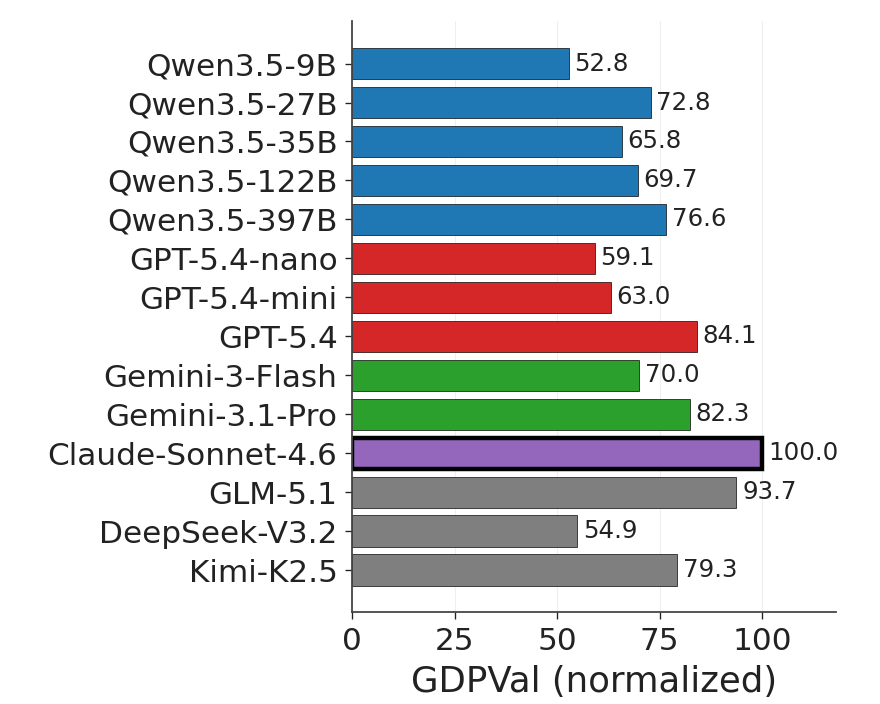}
  \includegraphics[width=0.55\linewidth]{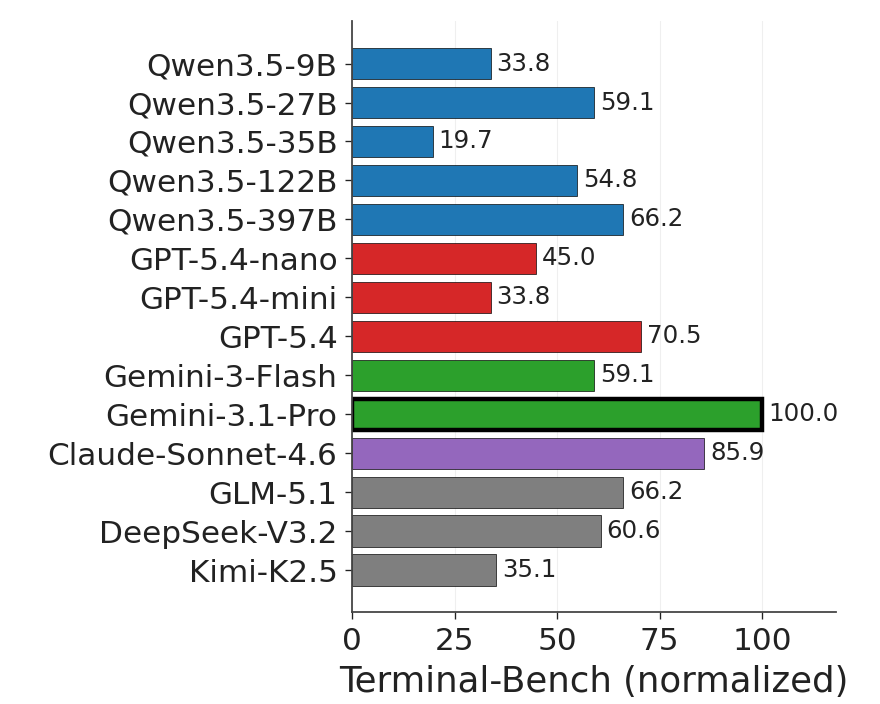}
  \centering
  \includegraphics[width=0.55\linewidth]{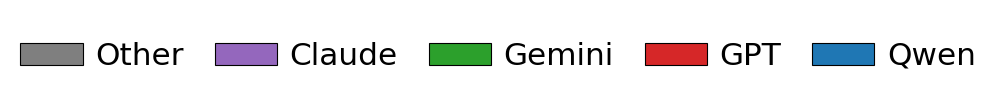}
  \caption{Zero-shot and full-specification performance. \textbf{Top:} Shopping-MMLU query product ranking mean NDCG over ${\sim}150$ queries per model, with $\pm$SEM. \textbf{Middle:} GDPVal normalized AA-leaderboard Elo (best$=100$). \textbf{Bottom:} Terminal-Bench normalized AA-leaderboard score (best$=100$).}
  \label{fig:zero-shot}
\end{figure}

We report the full per-model breakdowns behind the statistics in Section~\ref{subsec:setups}: capability baseline, behavioral decomposition, commit-turn/rank scatter, cumulative information gain, CommitAcc@1 by intent tier, per-cell $\mathrm{TotalIG}$, and the four-quadrant performance/uncertainty distribution.
\paragraph{Task-capability baseline.} Each model's zero-shot performance on the underlying benchmark, before any underspecification is introduced (Figure~\ref{fig:zero-shot}; Section~\ref{subsec:setups}). Shopping-MMLU is saturated across all $14$ models ($[81.1, 88.4]\%$ NDCG); GDPVal ($[52.8, 100]\%$) and Terminal-Bench ($[19.7, 100]\%$) span wide, giving the unsaturated regime we exploit in Section~\ref{subsec:cap-vs-align}.

\begin{figure}[!htbp]
  \centering
  \begin{subfigure}[b]{\linewidth}
    \centering
    \includegraphics[width=0.85\linewidth]{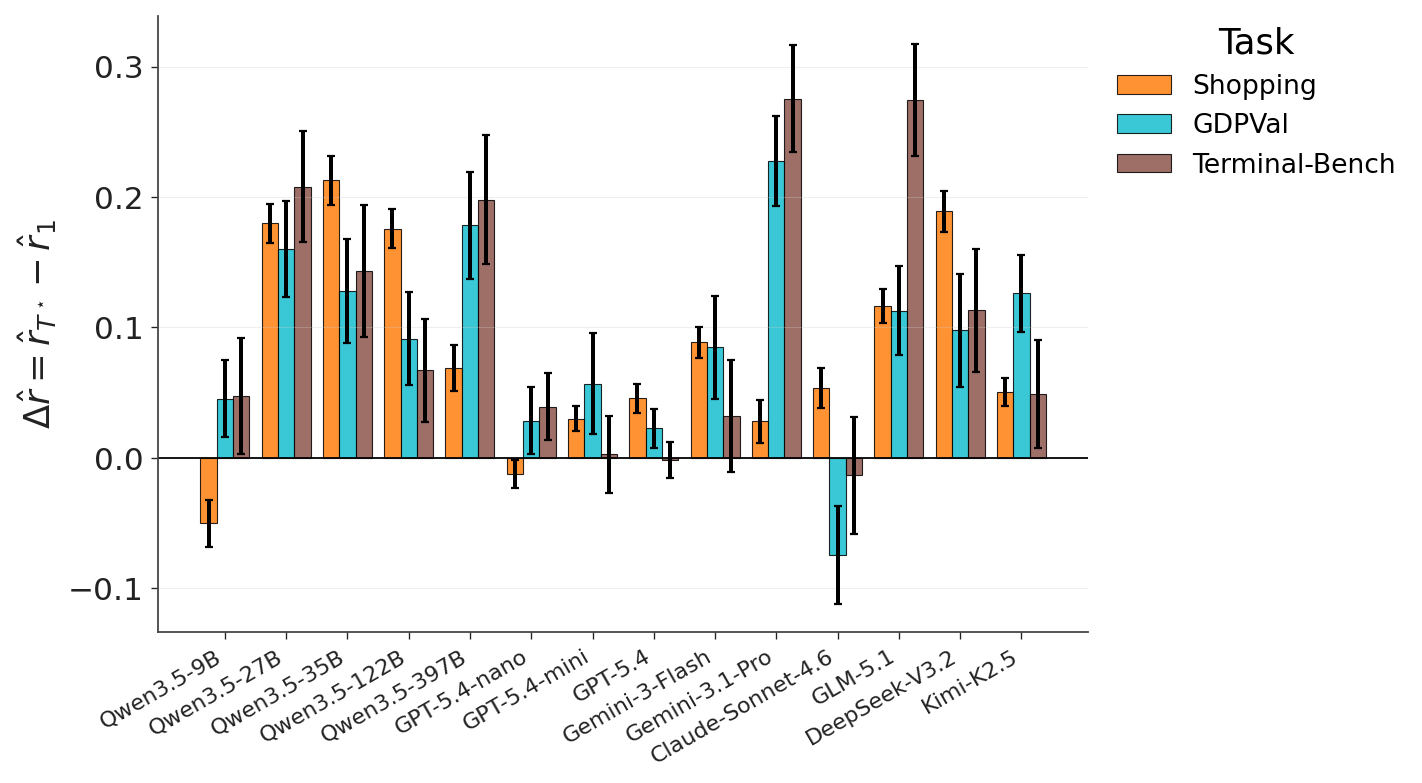}
    \caption{Rank improvement $\Delta\hat r = \hat r_{T^\star} - \hat r_1$ for all 14 evaluated models. Smaller x-axis font to fit the full set; otherwise identical to the main-text figure.}
    \label{fig:delta-rank-bar-full}
  \end{subfigure}\\[6pt]
  \begin{subfigure}[b]{\linewidth}
    \centering
    \includegraphics[width=0.7\linewidth]{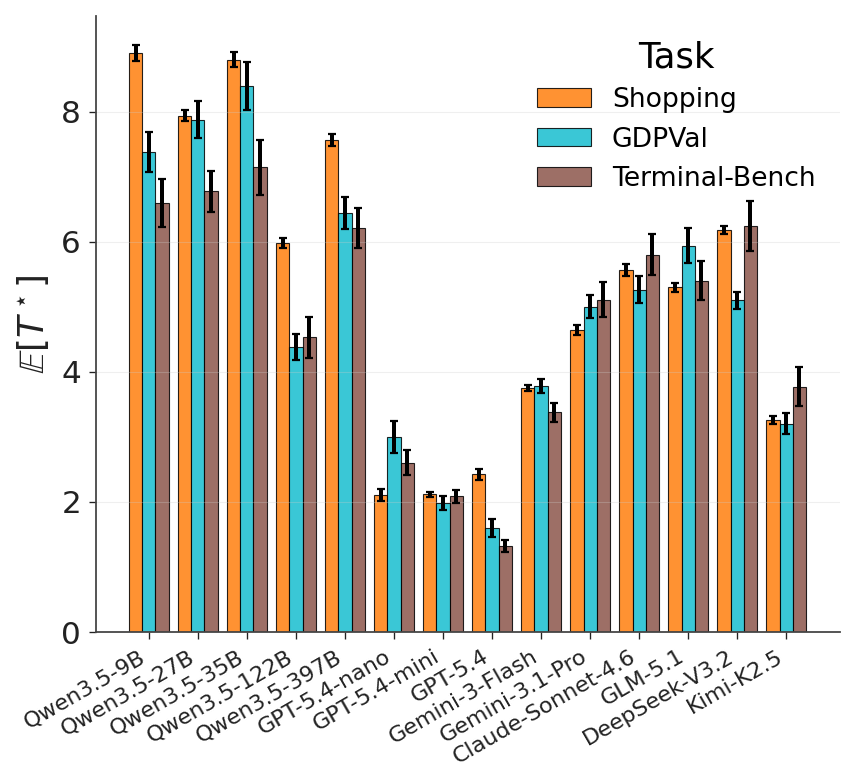}
    \caption{Mean commit turn $\mathbb{E}[T^\star]$ for all 14 evaluated models.}
    \label{fig:tstar-full}
  \end{subfigure}
  \caption{Per-model behavioral statistics for all 14 evaluated models.}
  \label{fig:behavioral-stats-full}
\end{figure}

\begin{figure}[!htbp]
  \centering
  \includegraphics[width=0.75\linewidth]{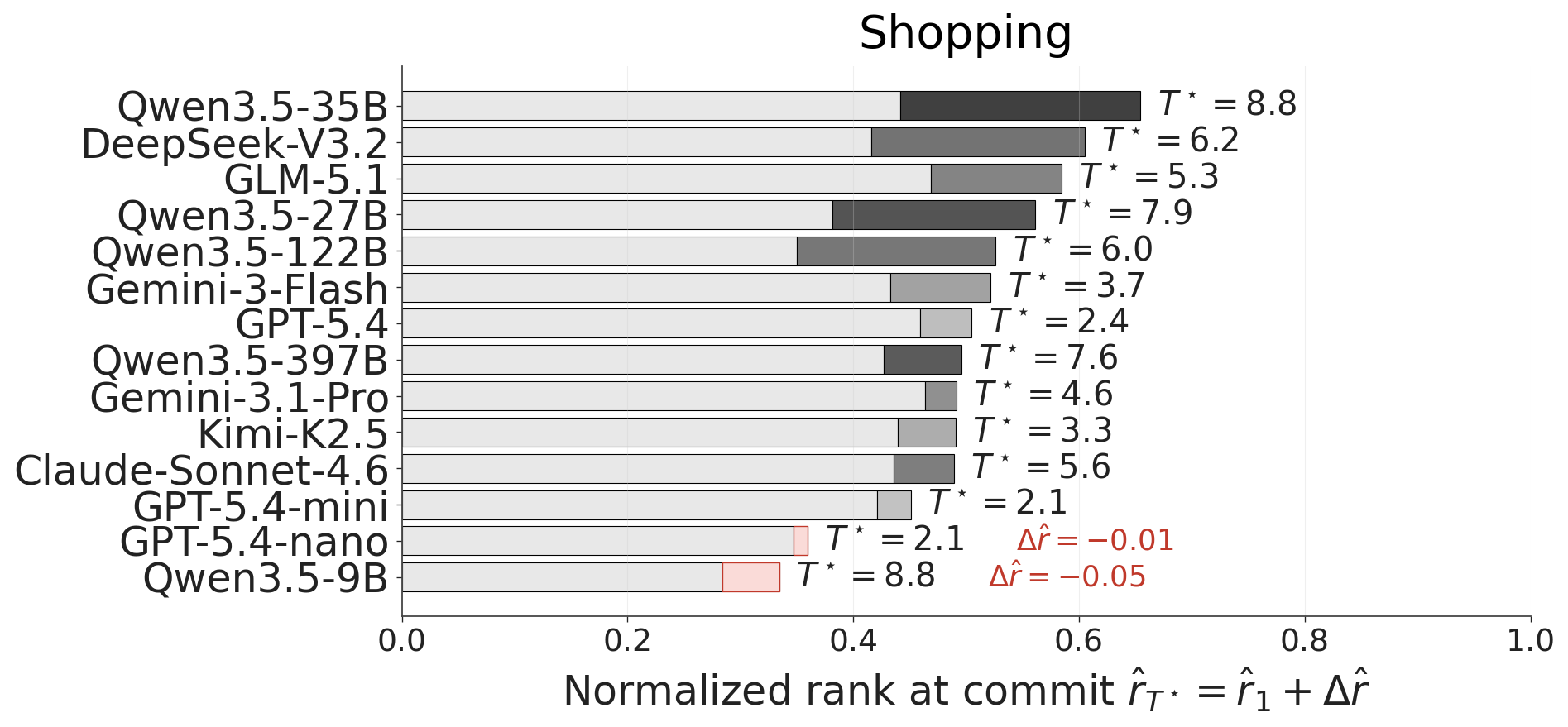}\\[2pt]
  \includegraphics[width=0.75\linewidth]{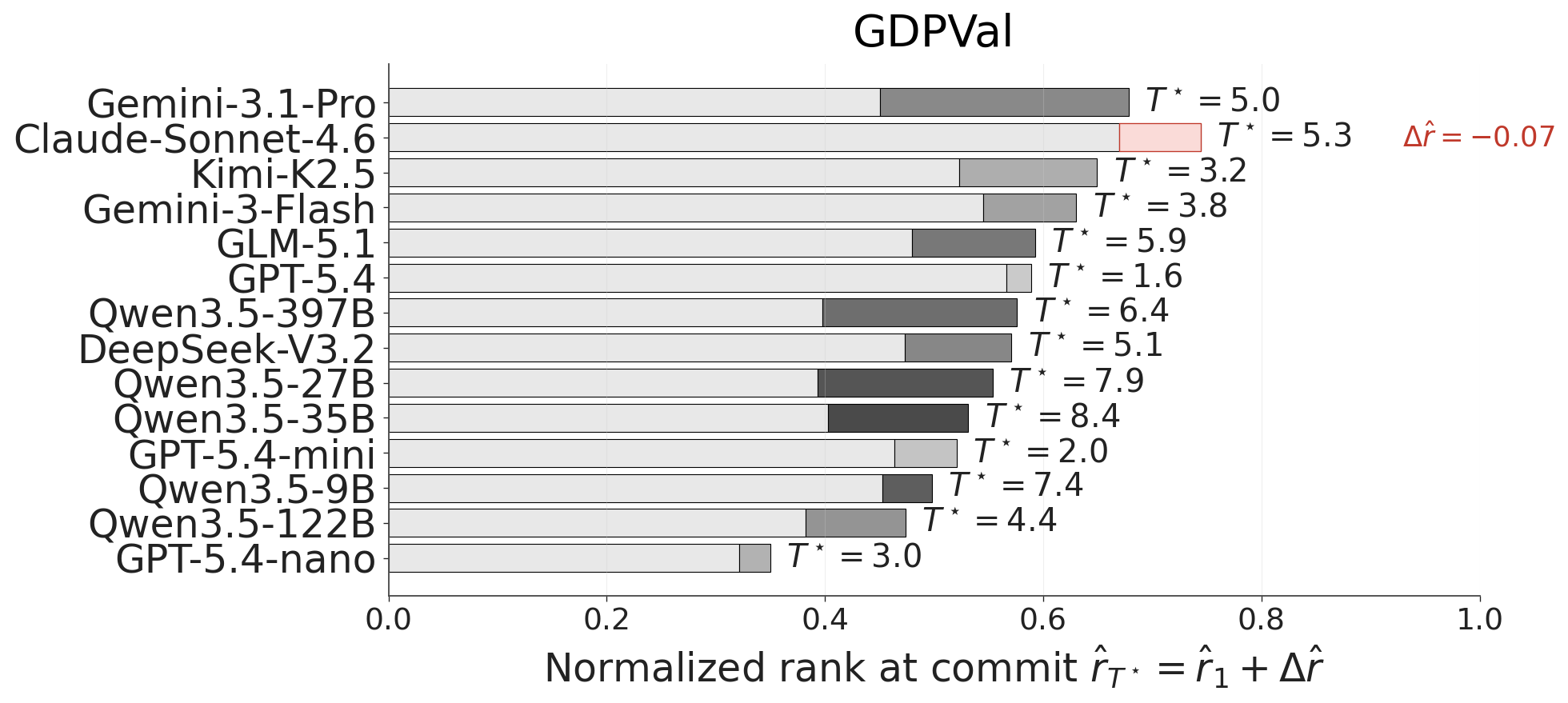}\\[2pt]
  \includegraphics[width=0.75\linewidth]{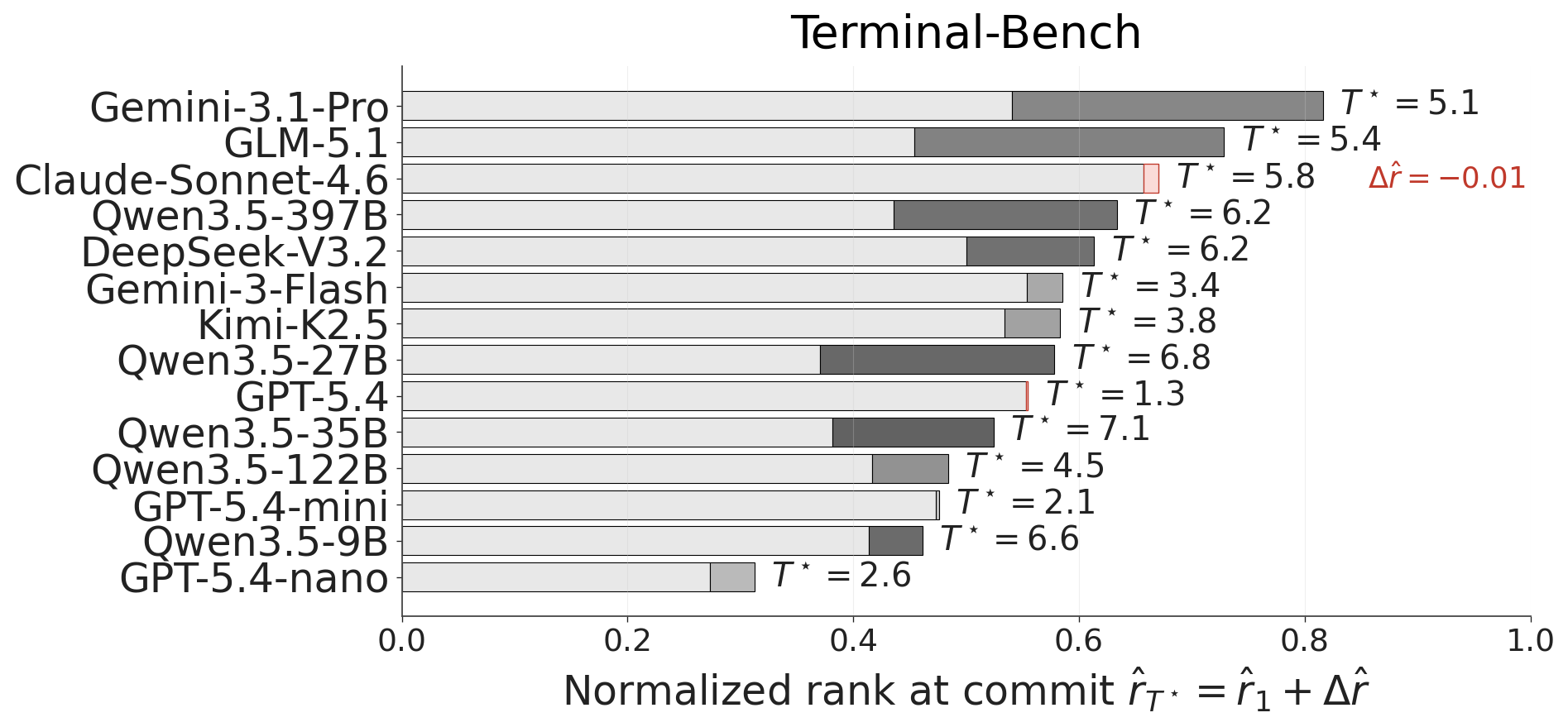}\\[2pt]
  \includegraphics[width=0.6\linewidth]{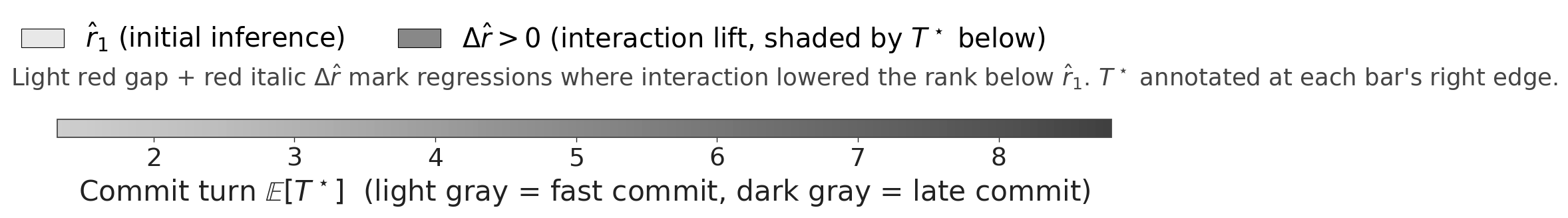}
  \caption{Behavioral decomposition along (1) the model's initial guess $\hat r_1$ (light gray), (2) how effectively the model interacts $\Delta \hat r = \hat r_{T^\star} - \hat r_1$ (dark gray), and (3) when the model commits $T^\star$ (depth of gray: light $=$ fast commit, dark $=$ late commit). The light red gap and red italic $\Delta \hat r$ mark regressions where interaction \emph{worsens} rank. One row per task: Shopping, GDPVal, Terminal-Bench.}
  \label{fig:behavioral-decomp-static}
\end{figure}

\begin{figure}[!htbp]
\centering
\includegraphics[width=0.32\linewidth]{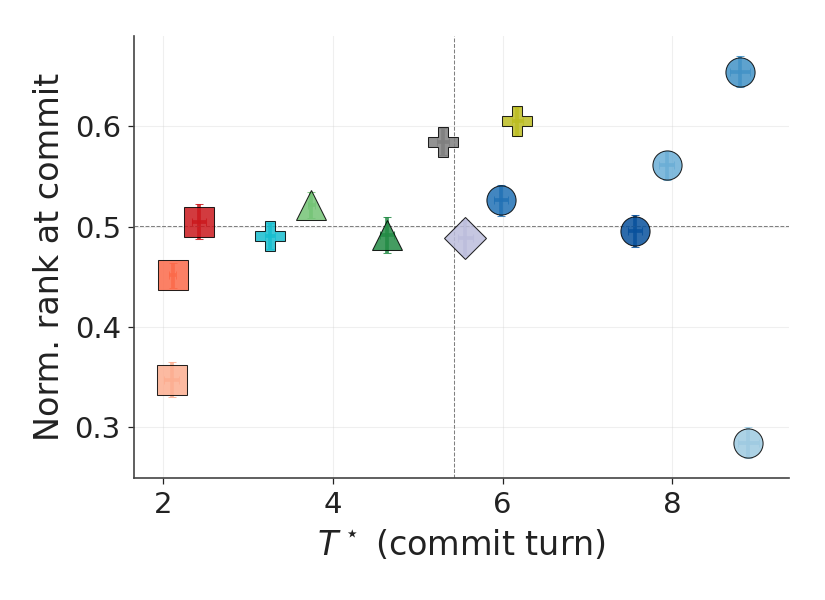}
\hfill
\includegraphics[width=0.32\linewidth]{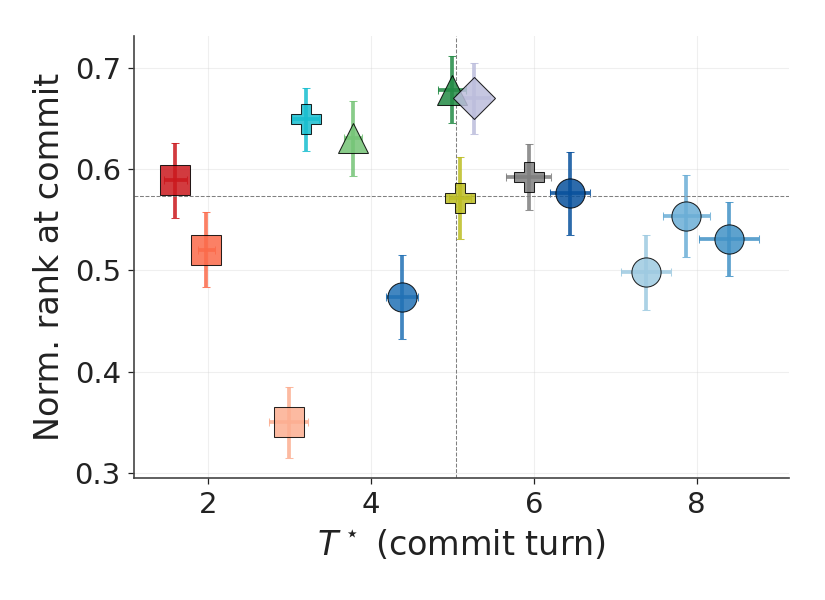}
\hfill
\includegraphics[width=0.32\linewidth]{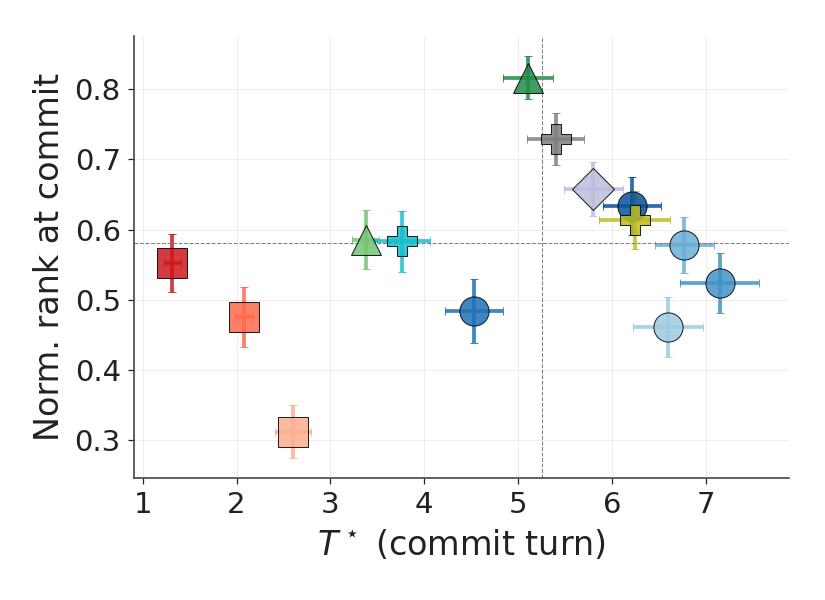}

\vspace{5pt}
\includegraphics[width=0.5\linewidth]{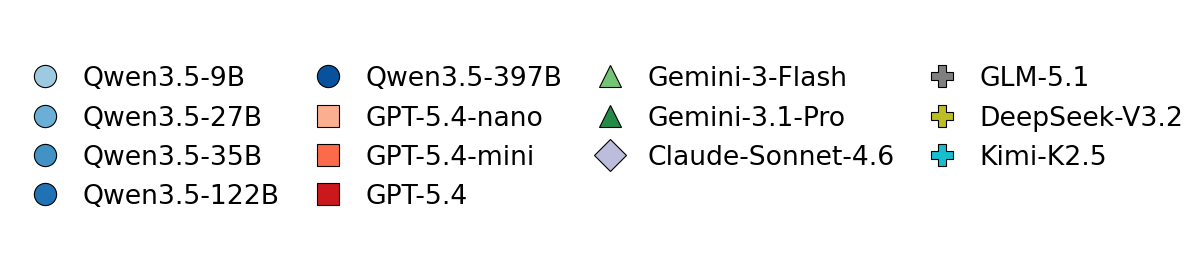}
\caption{Commit turn $T^\star$ vs.\ normalized rank at commit, per model (abstract tier). \textbf{Left:} Shopping. \textbf{Middle:} GDPVal. \textbf{Right:} Terminal-Bench.}
\label{fig:termination-rank}
\end{figure}

\paragraph{Behavioral statistics across all 14 models.} We report three complementary views of the $14$-model behavioral statistics at the abstract tier: 
\begin{enumerate}
    \item Rank improvement $\Delta\hat r = \hat r_{T^\star} - \hat r_1$ and commit timing $\mathbb{E}[T^\star]$ per model (Figure~\ref{fig:behavioral-stats-full}) extend Figure~\ref{fig:behavioral-decomp} to the full model set. $T^\star$ separates cleanly into three groups: early commit (GPT-5.4 family, $\le\!3$ turns), mid commit (Gemini and Claude, $4$--$6$), and late commit (Qwen3.5, $6$--$9$).

    \item The three-axis decomposition (Figure~\ref{fig:behavioral-decomp-static}) turns each rollout to per-model scalars: initial rank $\hat r_1$ (light gray), interaction lift $\Delta \hat r$ (dark gray), and commit turn $T^\star$ (shading depth of the delta bar: light $=$ fast, dark $=$ late); a light-red gap and a red italic $\Delta \hat r$ mark regressions where interaction \emph{worsens} rank. 

    \item The per-rollout $(T^\star, \hat r_{T^\star})$ scatter with per-model means (Figure~\ref{fig:termination-rank}) showcases the joint distribution of performance and commit timing: GPT-5.4 sits in the low-$T^\star$ / low-$\hat r_{T^\star}$ corner (fast commit, low rank), while the Qwen3.5 family occupies the high-$T^\star$ region with wider rank spread.
\end{enumerate}

\begin{figure}[!htbp]
  \centering
  \includegraphics[width=\linewidth]{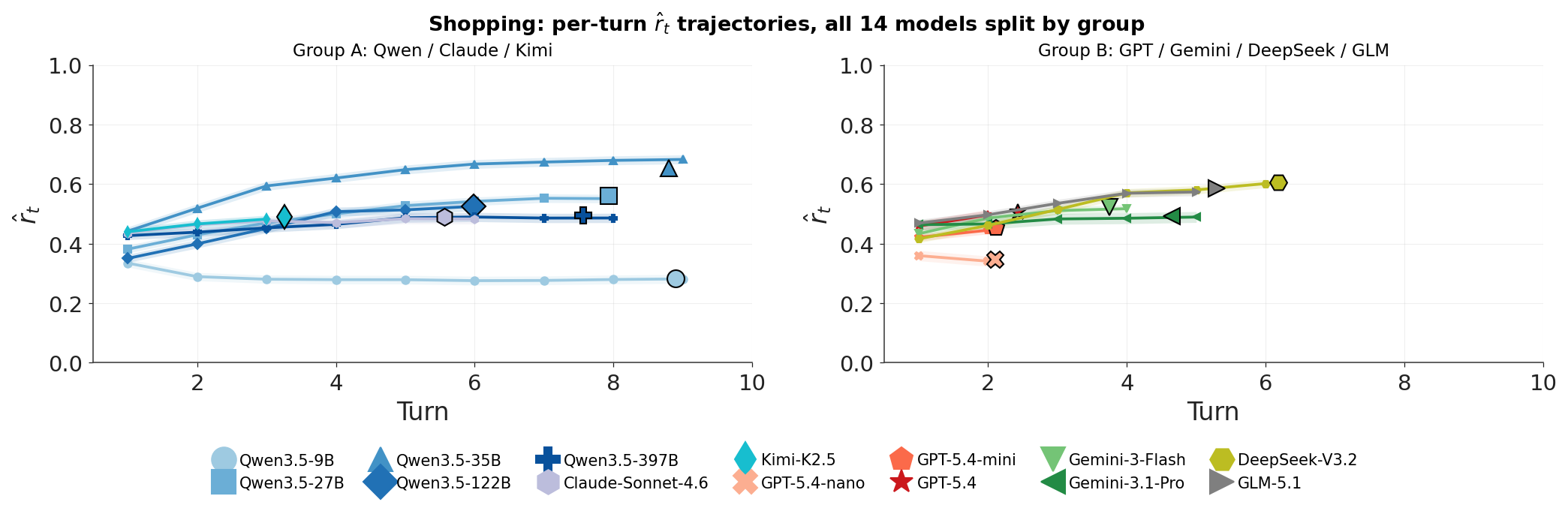}
  \caption{Shopping: per-turn trajectories for all 14 models, split into 2 groups of 7.}
  \label{fig:trajectory-full-shopping}
\end{figure}

\begin{figure}[!htbp]
  \centering
  \includegraphics[width=\linewidth]{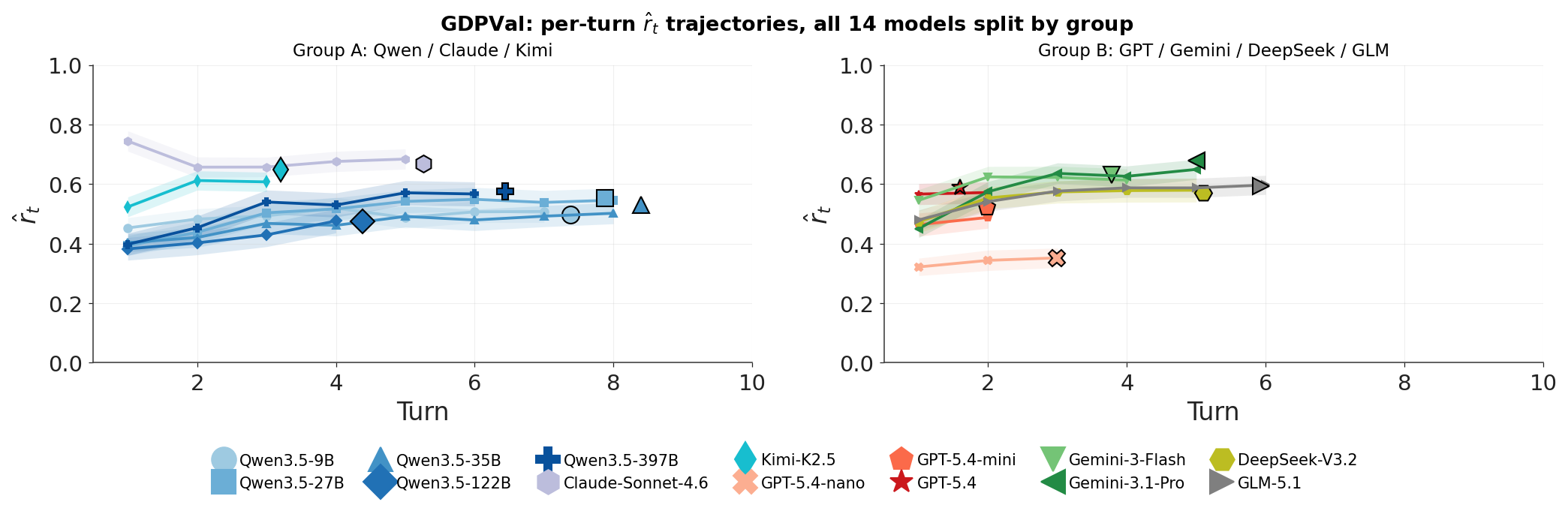}
  \caption{GDPVal: per-turn trajectories for all 14 models, split into 2 groups of 7.}
  \label{fig:trajectory-full-gdpval}
\end{figure}

\begin{figure}[!htbp]
  \centering
  \includegraphics[width=\linewidth]{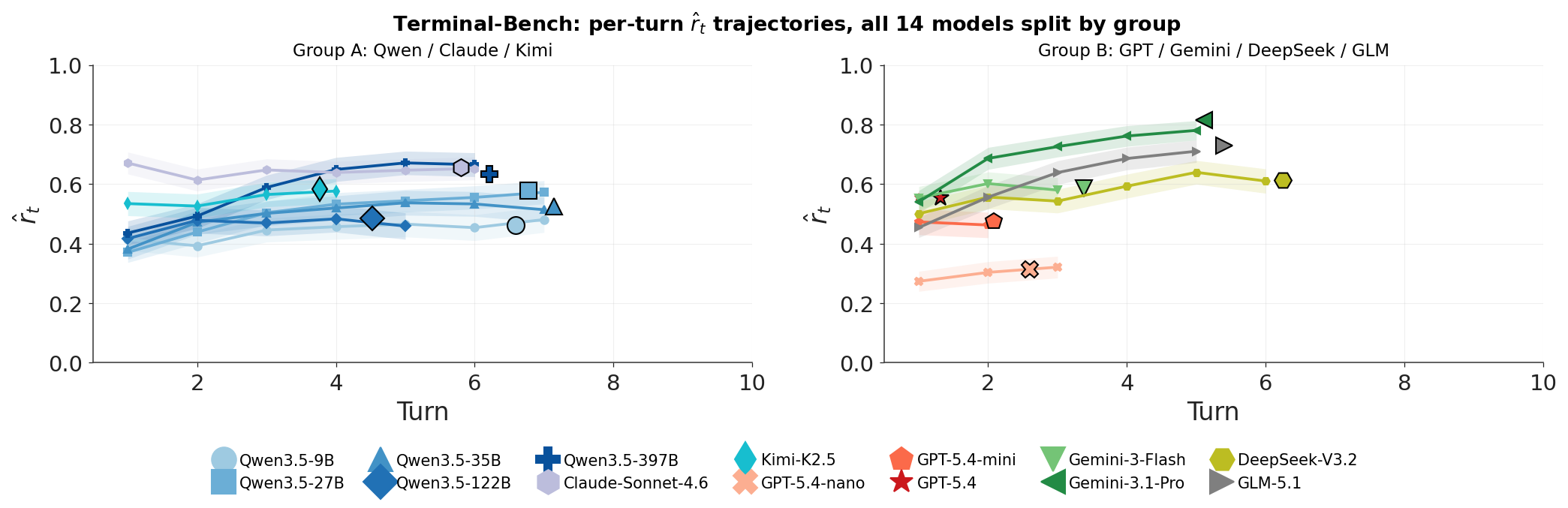}
  \caption{Terminal-Bench: per-turn trajectories for all 14 models, split into 2 groups of 7.}
  \label{fig:trajectory-full-terminalbench}
\end{figure}

\newpage
\paragraph{Full 14-model per-turn rank trajectories.} Figures~\ref{fig:trajectory-full-shopping}--\ref{fig:trajectory-full-terminalbench} extend the main-body per-turn trajectory (Figure~\ref{fig:behavioral-decomp}) to all $14$ evaluated models, split into two groups of seven per task so no panel becomes unreadable. \textbf{Group A} covers Qwen3.5 (5 sizes), Claude Sonnet-4.6, and Kimi-K2.5. \textbf{Group B} covers the GPT-5.4 family, Gemini-3-Flash and 3.1-Pro, DeepSeek-V3.2, and GLM-5.1. Only $\hat r_t$ is shown here; entropy $\tilde H_t$ is undefined for the $9$ models who do not emit confidence scores (i.e. logprobs) over $\mathcal{C}$, and its trajectory across the confidence-emitting subset is already in Figure~\ref{fig:behavioral-decomp}.

\begin{figure}[!htbp]
  \centering
  \includegraphics[width=0.85\linewidth]{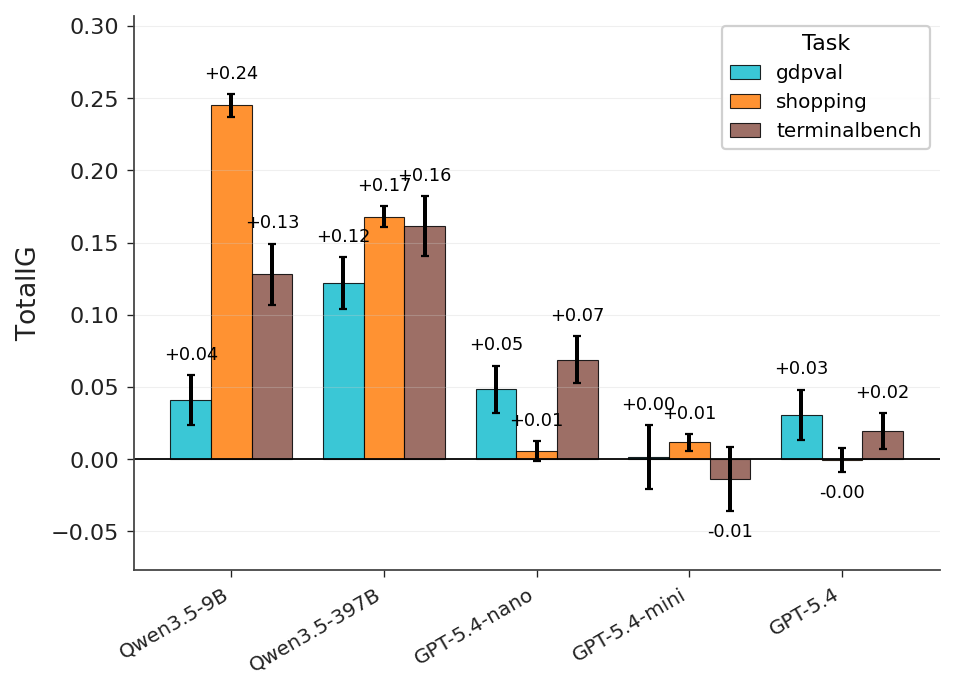}
  \caption{Cumulative information gain $\mathrm{TotalIG} = \tilde H_0 - \tilde H_{T^\star}$ per (model, task), averaged across the three intent tiers. Bars are mean $\pm$ pooled SEM across tiers.}
  \label{fig:total-ig-barplot}
\end{figure}

\paragraph{Cumulative information gain.} Per-(model, task) $\mathrm{TotalIG} = \tilde H_0 - \tilde H_{T^\star}$, averaged across the three intent tiers with pooled SEM (Figure~\ref{fig:total-ig-barplot}). Restricted to the five models whose ranker emits confidence scores over $\mathcal{C}$ (GPT-5.4 family, Qwen3.5-9B, Qwen3.5-397B-A17B). Qwen3.5 recovers substantial entropy on every task (tier-averaged $\mathrm{TotalIG}$ up to $0.24$ on Shopping and $0.16$ on Terminal-Bench); the GPT-5.4 family hovers near zero because $T^\star$ is too small to accumulate entropy reduction.

\newpage
\textbf{CommitAcc@1 by intent tier.} Per-model CommitAcc@1 across
abstract $\rightarrow$ moderate $\rightarrow$ concrete for each task
(Tables~\ref{tab:drift-shopping}--\ref{tab:drift-terminalbench}). The
\textsc{Average} row summarizes the alignment gap that closes as user
intent becomes more explicit: $22.0 \rightarrow 24.3 \rightarrow
27.9\%$ on Shopping, $27.6 \rightarrow 38.7 \rightarrow 88.0\%$ on
GDPVal, and $32.2 \rightarrow 41.5 \rightarrow 78.1\%$ on
Terminal-Bench.

\begin{table}[!htbp]
  \centering
  \small
  \caption{CommitAcc@1 (\%) across abstract / moderate / concrete intent tiers on Shopping.}
  \label{tab:drift-shopping}
  \begin{tabular}{l rrr}
    \toprule
    Model & Abs & Mod & Con \\
    \midrule
    qwen3.5-9b & $15.80 \pm 1.31$ & $19.25 \pm 1.42$ & $21.96 \pm 1.49$ \\
    qwen3.5-27b & $32.26 \pm 1.68$ & $32.78 \pm 1.68$ & $33.80 \pm 1.70$ \\
    qwen3.5-35b-a3b & $53.86 \pm 1.79$ & $44.79 \pm 1.79$ & $41.70 \pm 1.77$ \\
    qwen3.5-122b-a10b & $32.13 \pm 1.68$ & $27.89 \pm 1.61$ & $28.15 \pm 1.61$ \\
    qwen3.5-397b-a17b & $23.87 \pm 1.53$ & $23.90 \pm 1.54$ & $30.66 \pm 1.66$ \\
    \cmidrule(lr){1-4}
    gpt-5.4-nano & $7.97 \pm 1.37$ & $12.08 \pm 1.65$ & $17.22 \pm 1.92$ \\
    gpt-5.4-mini & $11.44 \pm 1.14$ & $16.97 \pm 1.35$ & $26.35 \pm 1.58$ \\
    gpt-5.4 & $12.89 \pm 1.70$ & $18.04 \pm 1.95$ & $24.68 \pm 2.19$ \\
    \cmidrule(lr){1-4}
    gemini-3-flash-preview & $18.51 \pm 1.39$ & $21.47 \pm 1.47$ & $27.76 \pm 1.61$ \\
    gemini-3.1-pro-preview & $13.88 \pm 1.76$ & $23.14 \pm 2.14$ & $30.33 \pm 2.33$ \\
    \cmidrule(lr){1-4}
    claude-sonnet-4-6 & $12.60 \pm 1.68$ & $22.37 \pm 2.12$ & $24.94 \pm 2.20$ \\
    \cmidrule(lr){1-4}
    glm-5.1 & $23.91 \pm 1.53$ & $26.09 \pm 1.58$ & $28.79 \pm 1.62$ \\
    deepseek-v3.2 & $32.52 \pm 1.68$ & $32.01 \pm 1.67$ & $31.11 \pm 1.66$ \\
    kimi-k2.5 & $16.45 \pm 1.33$ & $20.05 \pm 1.44$ & $23.39 \pm 1.52$ \\
    \midrule
    AVERAGE & $22.01 \pm 0.42$ & $24.34 \pm 0.45$ & $27.92 \pm 0.48$ \\
    \bottomrule
  \end{tabular}
\end{table}

\begin{table}[!htbp]
  \centering
  \small
  \caption{CommitAcc@1 (\%) across abstract / moderate / concrete intent tiers on GDPVal.}
  \label{tab:drift-gdpval}
  \begin{tabular}{l rrr}
    \toprule
    Model & Abs & Mod & Con \\
    \midrule
    qwen3.5-9b & $25.26 \pm 4.48$ & $35.79 \pm 4.94$ & $92.55 \pm 2.72$ \\
    qwen3.5-27b & $33.68 \pm 4.87$ & $46.32 \pm 5.14$ & $93.68 \pm 2.51$ \\
    qwen3.5-35b-a3b & $27.37 \pm 4.60$ & $37.89 \pm 5.00$ & $88.42 \pm 3.30$ \\
    qwen3.5-122b-a10b & $25.26 \pm 4.48$ & $33.68 \pm 4.87$ & $93.68 \pm 2.51$ \\
    qwen3.5-397b-a17b & $38.95 \pm 5.03$ & $49.47 \pm 5.16$ & $95.79 \pm 2.07$ \\
    \cmidrule(lr){1-4}
    gpt-5.4-nano & $9.47 \pm 3.02$ & $21.05 \pm 4.20$ & $65.26 \pm 4.91$ \\
    gpt-5.4-mini & $17.89 \pm 3.95$ & $17.89 \pm 3.95$ & $54.74 \pm 5.13$ \\
    gpt-5.4 & $22.11 \pm 4.28$ & $30.53 \pm 4.75$ & $69.47 \pm 4.75$ \\
    \cmidrule(lr){1-4}
    gemini-3-flash-preview & $33.68 \pm 4.87$ & $46.32 \pm 5.14$ & $96.84 \pm 1.80$ \\
    gemini-3.1-pro-preview & $40.00 \pm 5.05$ & $48.42 \pm 5.15$ & $97.89 \pm 1.48$ \\
    \cmidrule(lr){1-4}
    claude-sonnet-4-6 & $31.58 \pm 4.79$ & $49.47 \pm 5.16$ & $93.62 \pm 2.53$ \\
    \cmidrule(lr){1-4}
    glm-5.1 & $21.05 \pm 4.20$ & $45.26 \pm 5.13$ & $97.89 \pm 1.48$ \\
    deepseek-v3.2 & $31.58 \pm 4.79$ & $43.16 \pm 5.11$ & $94.74 \pm 2.30$ \\
    kimi-k2.5 & $28.42 \pm 4.65$ & $36.84 \pm 4.98$ & $96.81 \pm 1.82$ \\
    \midrule
    AVERAGE & $27.59 \pm 1.21$ & $38.72 \pm 1.32$ & $87.96 \pm 0.82$ \\
    \bottomrule
  \end{tabular}
\end{table}

\begin{table}[!htbp]
  \centering
  \small
  \caption{CommitAcc@1 (\%) across abstract / moderate / concrete intent tiers on Terminal-Bench.}
  \label{tab:drift-terminalbench}
  \begin{tabular}{l rrr}
    \toprule
    Model & Abs & Mod & Con \\
    \midrule
    qwen3.5-9b & $25.00 \pm 4.75$ & $34.94 \pm 5.27$ & $77.38 \pm 4.59$ \\
    qwen3.5-27b & $32.53 \pm 5.17$ & $45.12 \pm 5.53$ & $86.59 \pm 3.79$ \\
    qwen3.5-35b-a3b & $29.27 \pm 5.06$ & $39.02 \pm 5.42$ & $79.52 \pm 4.46$ \\
    qwen3.5-122b-a10b & $29.41 \pm 4.97$ & $38.55 \pm 5.37$ & $79.27 \pm 4.50$ \\
    qwen3.5-397b-a17b & $39.29 \pm 5.36$ & $54.12 \pm 5.44$ & $83.33 \pm 4.09$ \\
    \cmidrule(lr){1-4}
    gpt-5.4-nano & $7.06 \pm 2.79$ & $10.59 \pm 3.36$ & $50.59 \pm 5.46$ \\
    gpt-5.4-mini & $17.65 \pm 4.16$ & $25.88 \pm 4.78$ & $57.65 \pm 5.39$ \\
    gpt-5.4 & $24.71 \pm 4.71$ & $23.53 \pm 4.63$ & $62.35 \pm 5.29$ \\
    \cmidrule(lr){1-4}
    gemini-3-flash-preview & $34.12 \pm 5.17$ & $49.41 \pm 5.46$ & $78.82 \pm 4.46$ \\
    gemini-3.1-pro-preview & $61.18 \pm 5.32$ & $69.41 \pm 5.03$ & $91.76 \pm 3.00$ \\
    \cmidrule(lr){1-4}
    claude-sonnet-4-6 & $34.12 \pm 5.17$ & $54.12 \pm 5.44$ & $84.52 \pm 3.97$ \\
    \cmidrule(lr){1-4}
    glm-5.1 & $51.76 \pm 5.45$ & $53.57 \pm 5.47$ & $92.59 \pm 2.93$ \\
    deepseek-v3.2 & $34.94 \pm 5.27$ & $35.71 \pm 5.26$ & $83.33 \pm 4.09$ \\
    kimi-k2.5 & $29.41 \pm 4.97$ & $46.99 \pm 5.51$ & $85.37 \pm 3.93$ \\
    \midrule
    AVERAGE & $32.17 \pm 1.32$ & $41.50 \pm 1.38$ & $78.08 \pm 1.16$ \\
    \bottomrule
  \end{tabular}
\end{table}

\paragraph{Per-cell $\mathrm{TotalIG}$.} Cell-level values (Table~\ref{tab:totalig}) behind Figure~\ref{fig:total-ig-barplot}: $\mathrm{TotalIG}$ for every (model, tier, task) triple in the score-vector subset. The GPT-5.4 family sits at $\approx 0$ on Shopping because its low commit budget ($T^\star \le 3$) leaves belief entropy essentially unchanged from turn $1$ to commit, while Qwen3.5-9B ($\mathrm{TotalIG} \in [-0.03, 0.29]$) and Qwen3.5-397B-A17B ($[0.09, 0.21]$) both accumulate measurable reduction across the three tasks.

\begin{table}[!htbp]
  \centering
  \small
  \caption{Total information gain TotalIG $= \sum_t \Delta \tilde H_t$ per (model, tier), across all three tasks. Score-vector subset only (5 models: GPT-5.4 family + Qwen3.5-9B / 397B-A17B).}
  \label{tab:totalig}
  \begin{tabular}{ll rrr}
    \toprule
    Model & Tier & Shopping & GDPVal & Terminal-Bench \\
    \midrule
    Qwen3.5-9B & Abstract & $\phantom{-}0.29 \pm 0.01$ & $-0.03 \pm 0.03$ & $\phantom{-}0.12 \pm 0.03$ \\
    Qwen3.5-9B & Moderate & $\phantom{-}0.24 \pm 0.01$ & $\phantom{-}0.05 \pm 0.03$ & $\phantom{-}0.12 \pm 0.04$ \\
    Qwen3.5-9B & Concrete & $\phantom{-}0.21 \pm 0.01$ & $\phantom{-}0.11 \pm 0.03$ & $\phantom{-}0.14 \pm 0.03$ \\
    \cmidrule(lr){1-5}
    Qwen3.5-397B-A17B & Abstract & $\phantom{-}0.21 \pm 0.01$ & $\phantom{-}0.09 \pm 0.03$ & $\phantom{-}0.16 \pm 0.04$ \\
    Qwen3.5-397B-A17B & Moderate & $\phantom{-}0.17 \pm 0.01$ & $\phantom{-}0.14 \pm 0.04$ & $\phantom{-}0.21 \pm 0.04$ \\
    Qwen3.5-397B-A17B & Concrete & $\phantom{-}0.12 \pm 0.01$ & $\phantom{-}0.14 \pm 0.02$ & $\phantom{-}0.11 \pm 0.03$ \\
    \cmidrule(lr){1-5}
    GPT-5.4-nano & Abstract & $\phantom{-}0.00 \pm 0.01$ & $\phantom{-}0.02 \pm 0.03$ & $\phantom{-}0.07 \pm 0.02$ \\
    GPT-5.4-nano & Moderate & $-0.00 \pm 0.01$ & $\phantom{-}0.04 \pm 0.03$ & $\phantom{-}0.07 \pm 0.03$ \\
    GPT-5.4-nano & Concrete & $\phantom{-}0.02 \pm 0.01$ & $\phantom{-}0.08 \pm 0.03$ & $\phantom{-}0.07 \pm 0.03$ \\
    \cmidrule(lr){1-5}
    GPT-5.4-mini & Abstract & $\phantom{-}0.01 \pm 0.01$ & $\phantom{-}0.02 \pm 0.04$ & $-0.01 \pm 0.04$ \\
    GPT-5.4-mini & Moderate & $\phantom{-}0.02 \pm 0.01$ & $\phantom{-}0.02 \pm 0.05$ & $-0.01 \pm 0.04$ \\
    GPT-5.4-mini & Concrete & $\phantom{-}0.00 \pm 0.01$ & $-0.04 \pm 0.03$ & $-0.01 \pm 0.04$ \\
    \cmidrule(lr){1-5}
    GPT-5.4 & Abstract & $-0.00 \pm 0.02$ & $\phantom{-}0.01 \pm 0.03$ & $\phantom{-}0.05 \pm 0.03$ \\
    GPT-5.4 & Moderate & $\phantom{-}0.00 \pm 0.02$ & $\phantom{-}0.05 \pm 0.04$ & $-0.01 \pm 0.02$ \\
    GPT-5.4 & Concrete & $-0.00 \pm 0.01$ & $\phantom{-}0.03 \pm 0.02$ & $\phantom{-}0.01 \pm 0.01$ \\
    \bottomrule
  \end{tabular}
\end{table}

\paragraph{Quadrant distribution.} Per-model rollout share across the four correctness $\times$ confidence quadrants (Table~\ref{tab:quadrants}; quadrants defined in Table~\ref{tab:taxonomy}) at the abstract tier. Qwen3.5-397B-A17B carries the highest \emph{confidently-correct} share on every task ($21.1\%$ Shopping, $24.2\%$ GDPVal, $27.4\%$ Terminal-Bench), while GPT-5.4-mini concentrates in \emph{confidently-wrong} on GDPVal ($48.4\%$) and Terminal-Bench ($48.2\%$) despite committing in $\le 3$ turns.

\begin{table}[!htbp]
  \centering
  \small
  \caption{Per-(task, model) quadrant distribution at the abstract tier (\% of rollouts), score-vector subset. Bold = column max within each task. Quadrants defined as in Table~\ref{tab:taxonomy}.}
  \label{tab:quadrants}
  \begin{tabular}{@{}l l rrrr@{}}
    \toprule
     & Model & Confidently-correct & Lucky & Confidently-wrong & Honestly-uncertain \\
    \midrule
    \multicolumn{6}{l}{\textit{Shopping}} \\
    & Qwen3.5-9B & $9.3$ & $\mathbf{6.8}$ & $36.3$ & $47.7$ \\
    & Qwen3.5-397B-A17B & $\mathbf{21.1}$ & $4.6$ & $\mathbf{43.0}$ & $31.3$ \\
    & GPT-5.4-nano & $3.3$ & $4.6$ & $33.7$ & $\mathbf{58.4}$ \\
    & GPT-5.4-mini & $6.6$ & $4.9$ & $35.6$ & $53.0$ \\
    & GPT-5.4 & $10.6$ & $2.3$ & $42.8$ & $44.3$ \\
    \cmidrule(lr){1-6}
    \multicolumn{6}{l}{\textit{GDPVal}} \\
    & Qwen3.5-9B & $7.4$ & $\mathbf{17.9}$ & $7.4$ & $\mathbf{67.4}$ \\
    & Qwen3.5-397B-A17B & $\mathbf{24.2}$ & $14.7$ & $16.8$ & $44.2$ \\
    & GPT-5.4-nano & $6.3$ & $3.2$ & $15.8$ & $74.7$ \\
    & GPT-5.4-mini & $11.6$ & $6.3$ & $\mathbf{48.4}$ & $33.7$ \\
    & GPT-5.4 & $13.7$ & $8.4$ & $31.6$ & $46.3$ \\
    \cmidrule(lr){1-6}
    \multicolumn{6}{l}{\textit{Terminal-Bench}} \\
    & Qwen3.5-9B & $10.7$ & $14.3$ & $4.8$ & $\mathbf{70.2}$ \\
    & Qwen3.5-397B-A17B & $\mathbf{27.4}$ & $11.9$ & $17.9$ & $42.9$ \\
    & GPT-5.4-nano & $4.7$ & $2.4$ & $16.5$ & $76.5$ \\
    & GPT-5.4-mini & $7.1$ & $10.6$ & $\mathbf{48.2}$ & $34.1$ \\
    & GPT-5.4 & $9.4$ & $\mathbf{15.3}$ & $30.6$ & $44.7$ \\
    \bottomrule
  \end{tabular}
\end{table}

\newpage

\newpage

\newpage
\section{Training details}
\label{sec:app-model-training}

We fine-tune Qwen3.5-9B using two methods on Shopping with a user initialized at concrete intent tier: \textbf{SFT} (selected from BoN sampling $n=5$) and \textbf{RL} (GRPO from the base checkpoint on CommitReward). Following the convention of lowest reasoning setting, both training runs patch the base to a no-think variant, so the model does not emit \texttt{<think>} traces during training or evaluation.

\begin{table}[!htbp]
\begin{minipage}[t]{0.46\linewidth}
  \centering
  \small
  \setlength{\tabcolsep}{4pt}
  \caption{SFT hyperparameters. Full-parameter fine-tuning with cross-entropy on demonstration transcripts.}
  \label{tab:sft-config}
  \begin{tabular}{ll}
    \toprule
    Base           & Qwen3.5-9B (no-think) \\
    Train data     & $n=500$ transcripts \\
    Learning rate  & $2\times 10^{-5}$ \\
    Schedule       & cosine, $5\%$ warmup \\
    Grad clip      & $1.0$ \\
    Batch size     & $16$ \\
    Epochs         & $10$ \\
    Sequence len.\ & $8192$ \\
    Hardware       & $2\times$ H100 \\
    \bottomrule
  \end{tabular}
\end{minipage}\hfill
\begin{minipage}[t]{0.50\linewidth}
  \centering
  \small
  \setlength{\tabcolsep}{4pt}
  \caption{GRPO hyperparameters. Full-parameter RL from the base checkpoint (no SFT init); terminal reward is $\hat r_{T^\star}$ (terminal bonus $1.0$, truncation penalty $-0.5$ at the 15-turn interaction cap, step penalty $0$).}
  \label{tab:grpo-config}
  \begin{tabular}{ll}
    \toprule
    Base                & Qwen3.5-9B (no-think) \\
    Train / val         & $200$ / $16$ prompts \\
    Learning rate       & $5\times 10^{-7}$ \\
    Batch size          & $8$ prompts \\
    Total epochs (steps) & $20$ ($\sim\!90$) \\
    Group size ($n$)    & $8$ rollouts / prompt \\
    Repeats per prompt  & $4$ \\
    Sampling temp.\     & $1.0$ \\
    Multi-turn cap      & $15$ user $/ 15$ assistant \\
    Entropy coef.\      & $0.001$ \\
    KL regularizer      & in actor loss, coef $0.04$ \\
    Rollout engine      & vLLM (async), response $8192$ \\
    Hardware            & $2\times$ H100 \\
    \bottomrule
  \end{tabular}
\end{minipage}
\end{table}

\paragraph{Reward.} During RL the terminal signal is the normalized rank at commit, $\hat r_{T^\star}$, with terminal bonus $1.0$, $-0.5$ truncation penalty if the model reaches 15-turns without committing, and no step penalty ($\alpha=1.0$, $\lambda=0$ during training: the main paper's post hoc CommitReward $\alpha=1.5$, $\lambda=0.02$ is applied only at evaluation). The reward ranker is a separate frozen model (Qwen3.5 9B called via API), not updated during RL and sharing no weights with the policy, so the reward function is stationary throughout training. GRPO standardizes rewards within each $n\!=\!8$ rollout group before the policy-gradient update. The KL term to the base policy is applied as an actor-loss regularizer with coefficient $0.04$ and a low-variance estimator.

\paragraph{Rollouts.} Each RL rollout is a full task alignment episode against the LLM user simulator (Section~\ref{subsec:framework}), capped at $15$ turns per side and terminated when the assistant emits its commit signal. We repeat each prompt $4$ times per step to reduce variance from the user simulator.

\paragraph{Selection.} We select the best checkpoint by held-out $\hat r_{T^\star}$ on the $16$-item validation set, sampled every $5$ steps; the reported RL model is step $90$.

\begin{figure}[!htbp]
  \centering
  \tiny
  \includegraphics[width=0.5\linewidth]{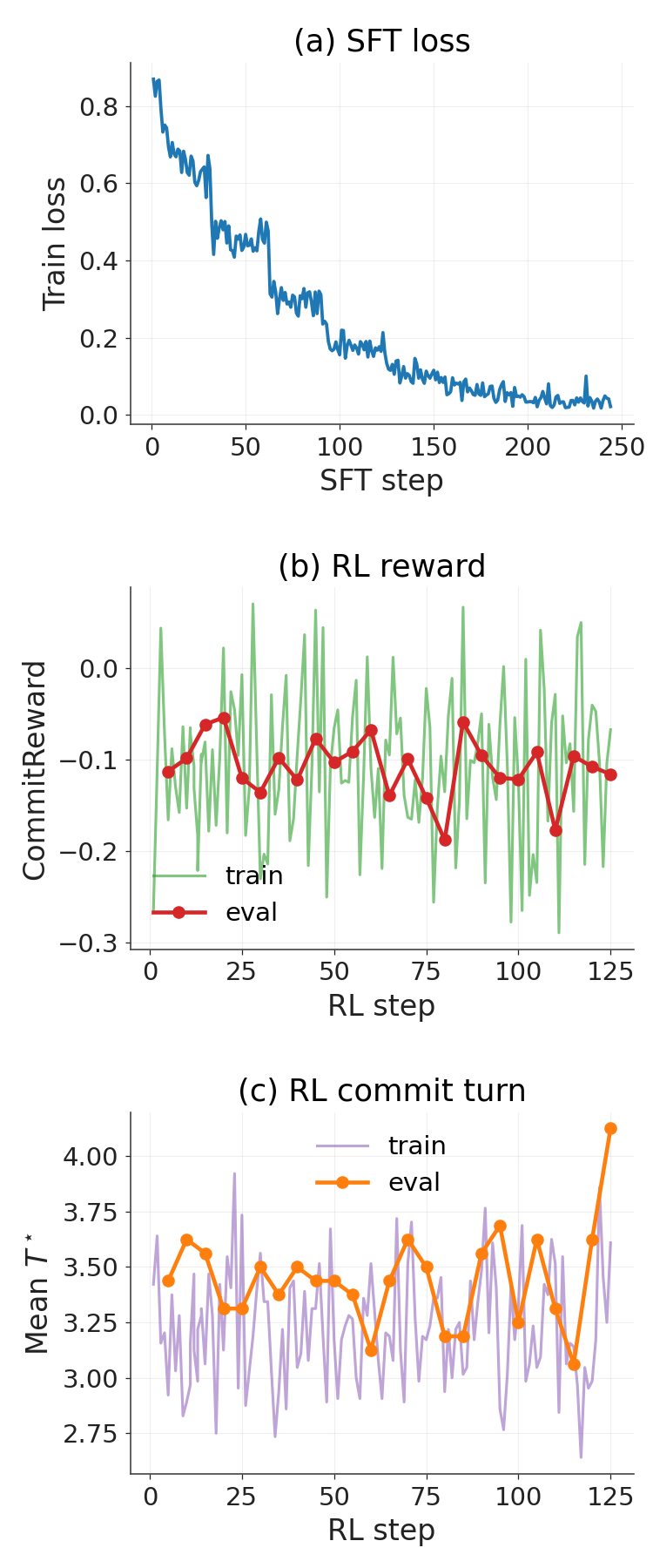}
  \caption{Qwen3.5-9B training curves, stacked top-to-bottom. \textbf{(a)} SFT loss. \textbf{(b)} RL train/eval CommitReward. \textbf{(c)} RL train/eval mean $T^\star$.}
  \label{fig:training-curves}
\end{figure}

\begin{figure}[!htbp]
  \centering
  \tiny
  \includegraphics[width=\linewidth]{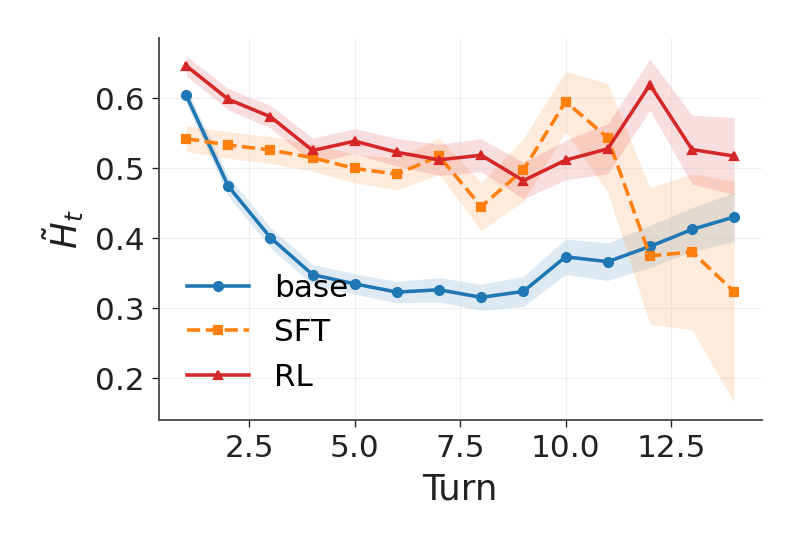}
  \caption{Qwen3.5-9B per-turn normalized entropy $\tilde H_t$ on Shopping abstract for base, SFT, and RL.}
  \label{fig:training-entropy}
\end{figure}

\newpage
\section{Prompts}
\label{sec:app-prompts}

Task-specific templates (ranker, user simulator, candidate ranking, intent evolver, candidate generation) are shown here in their Shopping form. GDPVal and Terminal-Bench variants have the same structure and only substitute domain nouns (``product'' $\to$ ``professional task specification'' / ``terminal-task instruction''), so we do not repeat them here.

\paragraph{Assistant system prompt.} We base our assistant system prompt off the CollabLLM assistant system prompt~\citep{wu2025collabllmpassiverespondersactive}, both at benchmark evaluation and during SFT/RL training. It instructs the assistant to interpret user intent proactively, ask targeted clarifying questions rather than assume, and commit to task execution by appending the token \texttt{[[TERMINATE]]} once it is confident.
\begin{lstlisting}[style=promptstyle]
The assistant is designed to be helpful, proactive, and highly interactive.

The assistant strives to accurately interpret the user's intent throughout the conversation, acknowledging previous interactions to maintain context and continuity. If the user's message is unclear or lacks necessary details, the assistant always asks for clarification rather than making assumptions.

The assistant asks specific follow-up questions and offers suggestions based on the user's needs, avoiding vague or generic prompts. It proactively provides guidance and potential next steps.

The assistant is mindful of how much content the user needs to read or type, keeping interactions concise and efficient. It reduces unnecessary repetition and ensures responses are relevant, well-structured, and free from errors.

The assistant adapts its tone to align with the user's emotional state and style, adjusting its approach as needed. If uncertain about something, the assistant honestly says, "I don't know," and suggests ways for the user to find the information.

## Completing the task
Once the assistant is confident it has enough information to complete the task on the user's behalf, it commits to the answer and ends the conversation by appending the token [[TERMINATE]] to its message. Example:

    "Based on what you've told me, I'll go with <chosen option>. [[TERMINATE]]"

Until confident, the assistant keeps asking short, targeted clarifying questions -- one per turn -- to narrow down the user's intent.
\end{lstlisting}

\paragraph{User simulator.} The simulator role-plays a shopper with an evolving intent state (Section~\ref{subsec:framework}), constrained to answer only what the assistant asks and only from the current intent, so the assistant must drive the interaction:
\begin{lstlisting}[style=promptstyle]
You are role-playing as a real shopper talking with an AI shopping assistant. You have a current sense of what you want -- it may be abstract (a problem or vague desire) or specific (closer to a concrete product) -- and it evolves as you talk with the assistant.

## Your Current Intent:
{current_intent}

## Conversation History:
{conversation_history}

## Guidelines:
- Stay in Character: Role-play as a human shopper. You are NOT an AI. Maintain a consistent personality.
- Minimize Effort: Give brief, natural answers (1-2 sentences). Don't volunteer extra details -- let the assistant ask follow-up questions.
- Realistic Behavior: Occasionally be slightly vague or imprecise, like a real person who doesn't memorize every product spec.
- Stay Goal-Oriented: Answer only what is asked. No small talk, no digressions, no asking the assistant questions back, no requesting links or purchase steps.
- Answer From Your Intent: Base your answers only on what your current intent expresses. Do not invent specifics (brands, dimensions, materials) that your intent does not already include -- if the assistant asks about something outside your intent, it is fine to say you're not sure or have no preference yet.
- Yes/No Questions: Start with a clear "Yes" or "No" followed by at most one brief clarifying sentence.

## Output Format:
Output a JSON object with three entries:
- "current_answer" (str): Briefly summarize which product(s) the AI is currently recommending or leaning towards.
- "thought" (str): Your internal reasoning as the user -- what the assistant just asked, what aspect of your intent is most relevant, and how to answer concisely.
- "response" (str): Your actual reply to the assistant. Keep it to 1-2 sentences.

## Important Notes:
- Answer every turn. Do not try to end the conversation yourself -- the assistant will decide when it has enough information.
- Double check that the JSON object is formatted correctly with all three fields present.
\end{lstlisting}

\paragraph{Intent evolver.} At every turn the evolver rewrites the user's current intent based on the latest exchange, moving it toward $x^\star$ only when the conversation has plausibly surfaced the relevant detail:
\begin{lstlisting}[style=promptstyle]
You are simulating how a real user's intent evolves as they talk with an AI assistant.

The user began with an abstract need (a problem, a feeling, a vague desire) and is gradually forming a clearer picture of what they want as the conversation progresses. Each turn, their mental picture should reflect what the conversation has actually surfaced -- new preferences, constraints, categories, or ideas. If the conversation has taken them close to a specific solution, the intent is allowed to become very concrete, including matching the ground-truth target when that is where the dialogue naturally leads.

## Ground-truth target the user will eventually converge on (hidden from the simulator, shown to you only so you can steer the evolution):
Target: {item}
Description: {item_description}

## Current intent (before this update):
{current_intent}

## Conversation so far (from the user's POV):
{conversation_history}

## Your Task:
Write the user's intent AFTER the latest exchange. Let it move as far toward the ground-truth target as the conversation justifies -- small refinement if little was learned, a big jump if the assistant surfaced ideas that unlock the target. If the last turn added no new information, keep the intent unchanged.

Guidelines:
- The intent is a short first-person utterance (1-2 sentences), written as the user would say it.
- Only incorporate details the conversation has plausibly surfaced. Do not invent ground-truth details out of nowhere -- they must be anchored in the dialogue.
- It is okay (and desirable) for the intent to reach the ground-truth target once the conversation has naturally led there.

## Output Format (JSON):
{"reasoning": "<1-2 sentences on what changed and why, or why nothing changed>",
 "intent": "<the user's updated intent as a short first-person utterance>"}
\end{lstlisting}

\paragraph{Ranker.} After every assistant turn, the ranker is called in a separate thread with the conversation so far and orders the candidate set from most- to least-likely $x^\star$. The rank of $x^\star$ under this ordering feeds $\hat r_t$, $\hat r_{T^\star}$, and CommitAcc@1:
\begin{lstlisting}[style=promptstyle]
Based on the following conversation, determine which candidate item best matches the user's preferences and intent.

Conversation so far:
{chat_history}

Candidate items (JSON format):
{candidates_block}

Re-order the candidate items from most likely to least likely to be the item the user is looking for, based on the preferences and requirements expressed in the conversation.

Output requirements:
- Return a single JSON object with exactly one key: "candidates".
- The value of "candidates" MUST be a JSON array (token '[' ... ']'), NOT a string. Do not surround the array with quotes.
- Return MINIFIED JSON only (one line, no code fences, no prose, no comments).
- Start the response with {"candidates": and end with ]}.
- Do not include any other keys, wrappers, or text before/after the JSON.
- ONLY return the JSON object, do NOT return anything besides the JSON object.
\end{lstlisting}

\paragraph{Candidate ranking task.} The entropy probe (Appendix~\ref{section:entropy-calculation}) runs after each turn with the candidates assigned to freshly shuffled letters, and reads the model's letter logprobs at the first response position:
\begin{lstlisting}[style=promptstyle]
Conversation so far:
{conversation}

The user is trying to purchase one of these products:

{options_block}

Based on the conversation above, which product is the user trying to purchase? Answer with ONLY the letter.

Answer:
\end{lstlisting}

The null-calibration variant is identical except \texttt{\{conversation\}} is replaced with the placeholder \texttt{(no conversation context)}.

\paragraph{Candidate generation (Stages 1--3).} For GDPVal and Terminal-Bench (Shopping uses real Amazon search results directly), each row is generated with LLM calls to Claude Sonnet: one call for Stage 1 (axes + delta-sets), then 14 parallel calls for Stages 2--3 (one corrupt-and-repair call per distractor). Below we show the instructions used at each stage verbatim.

\paragraph{Stage 1: propose axes and delta-sets.} A single LLM call reads $x^\star$ and its rubric (or test code, for Terminal-Bench) and returns the axes plus 14 delta-sets as strict JSON.
\begin{lstlisting}[style=promptstyle]
You are generating a candidate set for a task-disambiguation benchmark. Read the ground-truth task x* below and its rubric (or test code), and propose the axes and delta-sets that will produce 14 plausible alternative tasks.

# GROUND-TRUTH TASK x*
{spec}

# RUBRIC / TESTS
{rubric_or_tests}

# STEP 1 -- Extract binding axes
An axis is a concrete value stated verbatim in x* that (i) can be changed by an in-place edit, and (ii) if changed would cause a specific rubric criterion or test assertion to fail. Axes must be atomic and independent -- bundle coupled values (a total and its parts, a name that encodes its subject) into ONE axis so no single-axis perturbation leaves the spec incoherent.

# STEP 2 -- Apply the MODALITY TEST with real judgment
For each axis, ask whether x*'s value is the single most-ordinary choice for that slot:
- FREEZE (perturbations: []) every axis where x*'s value is THE canonical default and any alternative is less ordinary -- standard ports, default extensions, conventional filenames, the primary tool/format/identity, and illustrative examples ("like Ctrl-C", "such as vim", "Hello world"). Frozen axes stay identical across all 15 candidates.
- PERTURB only multi-modal axes where several values are genuinely equally ordinary (a city, a person/entity name, an arbitrary file basename, an identifier). Draw alternatives ONLY from the equally-common pool.
- When unsure, FREEZE. It is fine if only a handful of axes vary.

# STEP 3 -- Latent-axis check
If a perturbed axis's ground-truth value is enforced by the rubric but never stated in x*'s instruction text, insert a natural sentence making it explicit -- otherwise distractors will carry constraints that x* does not visibly share, leaving x* uniquely unconstrained.

# STEP 4 -- Compose 14 delta-sets
Each delta-set assigns every multi-modal axis one perturbed value. Vary choices across delta-sets so no two are identical and each axis's alternatives are well covered.

# OUTPUT (strict JSON, no prose)
{"axes": [{"name": "...", "spec_quote": "...", "ground_truth": "...", "breaks": "...", "perturbations": [...]}],
 "distractors": [{"id": 1, "delta_set": {"axis_name": "perturbed_value", ...}, "rationale": "..."}]}
\end{lstlisting}

\paragraph{Stages 2--3: corrupt and repair (one call per distractor).} We fan out 14 parallel LLM calls, each given $x^\star$ and one delta-set from Stage 1. Each call applies the delta-set in place, propagates coupled changes, and then self-audits the result as a domain expert.
\begin{lstlisting}[style=promptstyle]
You are producing one distractor for a task-disambiguation benchmark. Take the ground-truth task x* below and apply the given delta-set, then self-audit the result. Return the final distractor text only.

# GROUND-TRUTH TASK x*
{spec}

# DELTA-SET
{delta_set}

# STEP 1 -- CORRUPT
Apply EVERY entry in the delta-set to x* in place. Preserve everything else verbatim -- structure, register, role opener, length (+/- 15

# STEP 2 -- REPAIR
Re-read the result as the domain expert who would actually perform this task. Find every internal contradiction or impossible combination and fix it -- WITHOUT reverting a listed perturbation, and WITHOUT drifting back toward x*. Catch things like a count that does not match the items it enumerates, a named framework paired with a different one's structure, an assessment scale mismatched to the stated subject (age, anatomy, condition), a timeline whose steps do not order consistently, or a deliverable format that contradicts its stated file type.

# OUTPUT (JSON only)
{"prompt": "<the final distractor as a single string>"}
\end{lstlisting}

\paragraph{Assembly.} After the 14 calls return, a deterministic Python step deduplicates the distractors by delta-set signature, inserts $x^\star$ at a seeded random position, and anonymizes all 15 entries to \texttt{task\_001}--\texttt{task\_015} before writing the row. No LLM call is involved.

\paragraph{Tiered intent seeds.} A fourth prompt generates the abstract, moderate, and concrete first-turn user messages that seed the user simulator. Anchor content rises with tier: abstract carries no deliverable category, moderate names the primary verb and one deliverable-format constraint, concrete restores the specific entity, deliverable basename, and key numeric parameters.
\begin{lstlisting}[style=promptstyle]
You are generating user-intent abstractions for a professional task specification, across three tiers of vagueness. Each should be first-person and sound like a real working professional -- natural and progressively imprecise, not a polished task brief.

Tier definitions:
- ABSTRACT (<=10 words): A raw goal, frustration, or pain point. NO deliverable category, NO entities, NO rubric criteria.
- MODERATE (<=25 words): The task framing. MUST include: the primary verb of the task prompt's first sentence, one domain noun, and one deliverable-format constraint (paraphrased). NO worksheet names, exact thresholds, or entity names.
- CONCRETE (1-3 sentences, ~150-300 chars): A specific situation described naturally. MUST include the exact deliverable basename, any named worksheet/section, the primary subject anchor (entity/regulation/period), and the key quantitative parameters -- verbatim from the prompt or rubric.

CRITICAL: If the prompt is intentionally vague about the deliverable, preserve that vagueness -- do not fabricate plausible specifics. Abstract and moderate must not copy verbatim phrases longer than 3 consecutive words from the task prompt.

Return ONLY a JSON array of exactly 3 strings: [abstract, moderate, concrete].

Occupation: {occupation}
Task prompt: {prompt}
\end{lstlisting}

\end{document}